\documentclass[sigconf, nonacm]{acmart}
\usepackage{graphicx}
\usepackage{algorithm}
\usepackage{algpseudocode}
\usepackage{amsmath}
\usepackage[font=small,labelfont=bf]{caption}
\usepackage{subcaption}
\usepackage[utf8]{inputenc}
\usepackage{amsmath}
\usepackage{amsthm}
\usepackage{scalerel}
\usepackage{bbm}

\usepackage{pgfplots}
\usepackage{pgfplotstable}
\usepgfplotslibrary{statistics}
\usepgfplotslibrary{patchplots}
\usepackage{csvsimple}
\usepackage{balance}

\DeclareMathOperator*{\argmax}{arg\,max}
\DeclareMathOperator*{\edgerv}{e_{\scaleto{RV}{4pt}}}
\DeclareMathOperator*{\edgevv}{e_{\scaleto{VV}{4pt}}}
\newtheorem{theorem}{Theorem}

\usepackage[textsize=tiny, textwidth=2.05in, colorinlistoftodos]{todonotes}

\usepackage{soul}
\usepackage[inline]{enumitem}

\setcopyright{none} % acmcopyright}
\copyrightyear{2022}
\acmYear{2022}
\acmDOI{} % 10.1145/1122445.1122456}

\acmISBN{} % 978-1-4503-XXXX-X/18/06}

\acmSubmissionID{123-A56-BU3}

\pgfplotsset{
/pgfplots/custom legend/.style={
legend image code/.code={
\draw [only marks,mark=square]
plot coordinates { 
(0.3cm,0cm)
};
}, },
}

\renewcommand\footnotetextcopyrightpermission[1]{}
\pgfplotsset{compat=1.17}

\usepackage{fancyhdr}

\begin{document}

% REMOVE THIS TO REVERT TO CAMERA READY
%\pagestyle{fancy}
%\lhead{Published in the Proceedings of the 13th ACM/IEEE International Conference on Cyber-Physical Systems (ICCPS 2022).}
\lhead{Accepted for publication at ICCPS 2022}
\rhead{}
%\lfoot{Accepted for publication at ICCPS 2022}
\cfoot{\thepage}

\sloppy

% \setlength{\marginparwidth}{2.05in}

\title{An Online Approach to Solve the Dynamic Vehicle Routing Problem with Stochastic Trip Requests for Paratransit Services}


\author{Michael Wilbur$^1$,
Salah Uddin Kadir$^2$, 
Youngseo Kim$^3$,
Geoffrey Pettet$^1$,
Ayan Mukhopadhyay$^1$,
Philip Pugliese$^4$,
Samitha Samaranayake$^3$,
Aron Laszka$^2$, and Abhishek Dubey$^1$
}

\affiliation{
$^1$Vanderbilt University, Nashville TN, United States\\
$^2$University of Houston, Houston TX, United States\\
$^3$Cornell University, Ithaca NY, United States\\
$^4$Chattanooga Area Regional Transportation Authority (CARTA), Chattanooga TN, United States}

%\author{Michael Wilbur}
%\affiliation{%
%  \institution{Vanderbilt University}
%  \city{Nashville, TN}
%  \country{United States}}
%\email{michael.p.wilbur@vanderbilt.edu}

%\author{Salah Uddin Kadir}
%\affiliation{%
%  \institution{University of Houston}
%  \city{Houston, TX}
%  \country{United States}}
  
%\author{Youngseo Kim}
%\affiliation{%
%  \institution{Cornell University}
%  \city{Ithaca, NY}
%  \country{United States}}
  
%\author{Geoff Pettet}
%\affiliation{%
%  \institution{Vanderbilt University}
%  \city{Nashville, TN}
%  \country{United States}}

%\author{Ayan Mukhopadhyay}
%\affiliation{%
%  \institution{Vanderbilt University}
%  \city{Nashville, TN}
%  \country{United States}}
  
%\author{Philip Pugliese}
%\affiliation{%
%  \institution{Chattanooga Area Regional Transportation Authority}
%  \city{Chattanooga, TN}
%  \country{United States}}
  
%\author{Samitha Samaranayake}
%\affiliation{%
%  \institution{Cornell University}
%  \city{Ithaca, NY}
%  \country{United States}}
  
%\author{Aron Laszka}
%\affiliation{%
%  \institution{University of Houston}
%  \city{Houston, TX}
%  \country{United States}}

%\author{Abhishek Dubey}
%\affiliation{%
%  \institution{Vanderbilt University}
%  \city{Nashville, TN}
%  \country{United States}}

\renewcommand{\shortauthors}{Wilbur, et al.}
%%

\begin{abstract}
  %Many transit agencies have to respond to trip requests that arrive in real-time. To avoid myopic decisions that lead to significant inefficiency in the long term, vehicles should be allocated to requests by optimizing some non-myopic utility function that considers potential future requests (e.g., the expected number of passengers served over an entire day). This entails solving hard combinatorial and sequential decision-making problems under uncertainty. To ensure scalability, recent work has focused on combining online routing-based heuristics (e.g., simulated annealing) with offline approaches (e.g., value function approximation). Another approach is to solve the problem in a myopic manner while considering a fixed batch of requests. We point out two major problems with such approaches. First, myopic approaches fail to account for future requests; this is particularly prominent for problem settings that are relatively sparse as fewer requests can be batched together. Second, in many problem settings, in many problem settings, the environment changes dynamically, which renders the offline approximations stale (e.g., traffic conditions and the number of available vehicles can change over time). As our collaboration with a major public transportation agency in the USA has revealed, a prototypical example in which both the aforementioned issues manifest is paratransit services. 
Many transit agencies operating paratransit and microtransit services have to respond to trip requests that arrive in real-time, which entails solving hard combinatorial and sequential decision-making problems under uncertainty.
To avoid decisions that lead to significant inefficiency in the long term, vehicles should be allocated to requests by optimizing a non-myopic utility function or by batching requests together and optimizing a myopic utility function. While the former approach is typically \textit{offline}, the latter can be performed \textit{online}. We point out two major issues with such approaches when applied to paratransit services in practice. First, it is difficult to batch paratransit requests together as they are temporally sparse. Second, the environment in which transit agencies operate changes dynamically (e.g., traffic conditions can change over time), causing the estimates that are learned offline to become stale. To address these challenges, we propose a fully online approach to solve the dynamic vehicle routing problem (DVRP) with time windows and stochastic trip requests that is robust to changing environmental dynamics by construction. We focus on scenarios where requests are relatively sparse---our problem is motivated by applications to paratransit services. We formulate DVRP as a Markov decision process and use Monte Carlo tree search to evaluate actions for any given state. Accounting for stochastic requests while optimizing a non-myopic utility function is computationally challenging; indeed, the action space for such a problem is intractably large in practice. To tackle the large action space, we leverage the structure of the problem to design heuristics that can sample promising actions for the tree search. Our experiments using real-world data from our partner agency show that the proposed approach outperforms existing state-of-the-art approaches both in terms of performance and robustness.

\end{abstract}

\keywords{Vehicle Routing Problem, Monte Carlo Tree Search, Online Planning, Smart Transit, Decision-making under Uncertainty}

\maketitle

\pagenumbering{arabic}

\section{Introduction}

The vehicle routing problem (VRP) is a well-known combinatorial optimization problem that seeks to assign a fleet of vehicles to routes to serve a set of customers/requests~\cite{dantzig1959truck}. Many real-world use cases of transportation agencies are modeled by the dynamic version of the problem (DVRP) with stochastic trip requests. In such settings, some customer requests may be known at the time of planning while others are unknown, and some stochastic information may be available about potential future requests~\cite{bent2004scenario}. Although the dynamic and the stochastic versions have traditionally been tackled separately, \citeauthor{bent2004scenario}~\cite{bent2004scenario} and \citeauthor{hvattum2006solving}~\cite{hvattum2006solving}, among others, showed that dynamic planning could use the stochastic information to improve performance. There are three broad approaches for solving DVRPs with stochastic requests. First, as requests arrive, a group of requests can be batched together, and routes can be optimized myopically for the particular batch in an \textit{online} manner~\cite{alonso2017demand}. Second, the routing problem can be solved to maximize a non-myopic utility function by learning a policy in an \textit{offline} manner that maps any given state of the problem to an action (i.e., a route plan for the vehicles)~\cite{nazari2018reinforcement}. Third, a combination of offline computation and online heuristics can be used for non-myopic planning~\cite{bent2004scenario,joe2020deep}.

A fundamental challenge in solving the DVRP non-myopically is computational tractability; indeed, real-world applications of DVRP are often intractable since they entail solving a hard combinatorial optimization problem with consideration for future requests. Prior work based on the combination of offline learning and online heuristics has addressed this bottleneck to some extent. \citeauthor{bent2004scenario}~\cite{bent2004scenario} continuously generate and store a pool of promising plans. Then, at the time of execution, they use a least-commitment strategy to select a key plan from the pool. \citeauthor{shah2020neural}~\cite{shah2020neural} use approximate dynamic programming and leverage a neural network-based approximation of the value function to handle the complexity from combinations of passenger requests. \citeauthor{joe2020deep}~\cite{joe2020deep} ensure near real-time response by combining online routing-based heuristics (e.g., simulated annealing) with offline approaches (e.g., value function approximation)~\cite{joe2020deep}. Myopic approaches, on the other hand, focus on pooling requests together to optimally allocate a specific batch of requests to routes~\cite{alonso2017demand}. 
A major challenge in such a setting is computing solutions fast enough to assign vehicles to requests in real-time (in practice, some delay is acceptable after a request is made). Hence, myopic approaches that operate in an online manner must be \textit{anytime} or able to compute a feasible action quickly; requests are stochastic and a decision must be computed before future requests arrive.  \citeauthor{alonso2017demand}~\cite{alonso2017demand} scale a myopic approach to assigning routes to pooled requests for real-world use cases by using an anytime algorithm that starts from a greedy solution and then improves it through constrained optimization.

%\Aron{transition between paragraph is rather abrupt here}
We focus on \textit{paratransit} services~\cite{lave2000state} in this paper, which are an important real-world example of DVRPs. Paratransit service is a socially beneficial curb-to-curb transportation service provided by public transit agencies for passengers who are unable to use fixed-route transit (e.g., passengers with disabilities). Our partner agency, the Chattanooga Area Regional Transportation Authority (CARTA), operates paratransit services in a mid-sized metropolitan area in the USA. While paratransit services resemble traditional on-demand ride-pooling services in some ways (e.g., the arrival of real-time requests and ride-sharing), our collaboration revealed some crucial differences between canonical examples of DVRPs in prior literature (e.g., ride-pooling or cargo delivery) and paratransit services. First, the frequency of requests is usually lower than services like taxis. For example, CARTA operates paratransit services in a metropolitan area with about 1.8 million people and usually serves about 200 requests per day (with 10 hours of operation each day). This constraint makes it difficult to batch requests together for myopic algorithms. Note that while the requests are temporally sparse, the decision for each request must be computed quickly. Second, paratransit services operate under the Americans with Disabilities Act (ADA), which enforces time windows as a hard constraint, unlike on-demand taxi services, thereby requiring strict adherence to such constraints. Our discussion with CARTA also revealed a potential issue with offline approaches for solving DVRPs. In practice, the environment in which CARTA operates is highly dynamic; traffic conditions in a city can change due to construction, events, or accidents, and the number of available vehicles or drivers can vary. In such cases, offline approximations can potentially lead to decisions that are far from optimal.

This paper introduces a \textit{fully online} approach for use-cases such as paratransit services that is \textit{anytime}, \textit{non-myopic}, \textit{robust} to dynamic changes in the environment, and also \textit{scalable} to real-world applications. Designing a completely online approach to solve DVRPs in a non-myopic setting is extremely challenging --- the action space for such a problem is intractably large for real-world applications. For example,  for our partner agency in a midsize metropolitan city in US with five vehicles, each with a capacity of eight passengers, the action space is of the order of $10^{22}$. 
%h just five vehicles with a capacity of , the action space is of the order of 
%$10^{22}$. 
Also, note that while requests are relatively sparse in our problem setting (one request every few minutes on average), computation for each decision needs to be fairly quick; naturally, it is infeasible to keep customers waiting for more than some exogenously defined duration.

The summary of contributions are as follows:
\begin{enumerate*}[label=\textbf{({\arabic*})}]
    \item  We design a fully online and non-myopic solver for DVRP with stochastic requests that scales to real-world problems by leveraging the structure of the problem instance. Our approach, MC-VRP (\textbf{M}onte \textbf{C}arlo tree search based solution for \textbf{v}ehicle \textbf{r}outing \textbf{p}roblem) is robust to environmental dynamics by construction.
    \item We model the DVRP  as a route-based Markov decision process (MDP)~\cite{ulmer2017route}. Given an arbitrary state of the MDP, we use generative models over customer requests and travel time to simulate the environment under consideration, which in turn enables us to use Monte Carlo tree search~\cite{kocsis2006bandit} to find promising actions for the state. Our approach does not require offline training, the only requirement is a generative model that can be sampled at run-time.
    \item To tackle the intractably large action space, we leverage the structure of the problem to find promising actions. Specifically, given a set of routes and a new request, we create a weighted graph based on a budget-based heuristic whose edges represent: (a) which vehicles can serve the new request, and (b) which vehicles can swap unpicked requests from their routes to maximize utility. Then, we sample independent sets from the graph that correspond to feasible actions for the given state of the MDP. 
   \end{enumerate*}
   %\Samitha{What is this nice structure?} 
   
To evaluate the proposed approach, we consider three baselines. First, we look at a greedy strategy, which assigns a new request to the vehicle that provides the highest myopic utility. Second, we look at recent work  by \citeauthor{joe2020deep}~\cite{joe2020deep}, which outperformed well-known state-of-the-art approaches such as the multiple scenario-based approach~\cite{bent2004scenario} and the approximate value iteration~\cite{ulmer2019preemptive}. Third, we compare our approach with a batched myopic setting by using the work by \citeauthor{alonso2017demand}~\cite{alonso2017demand}. Our experimental evaluations show that MC-VRP approach outperforms the baselines in terms of performance and robustness to changing environmental conditions.

\section{Problem Description and Model}

% In case the problem setting involves a combination of real-time and day-ahead requests (this is common in paratransit services),  the initial routes $\theta^i_{0}$ can be computed by using an offline VRP solver. 

%In our problem setting, passengers use an online service (e.g., a smartphone application or phone calls) to  book trips. 

% \usepackage{graphicx}
\begin{table}[t]
\footnotesize
\centering
\caption{Symbols}
\begin{tabular}{|l|l|}

\hline
\textbf{Notation} & \textbf{Description} \\ \hline
$V$ & A set of vehicles each of capacity $c$\\ \hline
$\theta^t$ & \begin{tabular}[c]{@{}l@{}}The set of route plans for all the vehicles with $\theta^i_t$ \\ denoting the plan vehicle $v_i \in V$ \\ \end{tabular} \\ \hline
$\Theta_t$ & The set of all possible route plans at time $t$\\ \hline
$vloc_t$ & \begin{tabular}[c]{@{}l@{}}A set consisting of locations of all vehicles at time $t$ \\ with $vloc^i_t$ denoting the location of $v_i \in V$\end{tabular}\\ \hline
$R_t$ & A new trip request at time $t$\\ \hline
$\mathcal{R}$ & Ordered set of trip requests for a day \\ \hline
$[e_t, p_t]$ & Service time window for request $R_t$ \\ \hline
$C(\theta^m_i)$ & Cost of vehicle $m$ servicing at route plan \\ \hline
$M$ & Set of vehicles in the paratransit fleet \\ \hline
$S_i$ & \begin{tabular}[c]{@{}l@{}}State tuple $<T_i, vloc_i, \theta_i>$ for\\ timestep $i$\end{tabular} \\ \hline
$E$ & Generative demand model \\ \hline
$K_{max}$ & \begin{tabular}[c]{@{}l@{}}Maximum number of feasible actions\\ to consider at each time epoch\end{tabular} \\ \hline
\end{tabular}%
\label{tab: symbols}
\end{table}

Table~\ref{tab: symbols} provides a notation lookup table used throughout this work. Typically, in paratransit services, passengers request pick-up times in advance of the trip, even when requesting on the same day. This is in contrast of on-demand ride services that are often requested with a short lead time. For example, customers can call in the morning to request for a ride during early afternoon. However, in some cases customers request rides with a short lead time. We assume that requests are i.i.d. according to a distribution $D$.
% In practice there is often enough gap between requests in paratransit services that we can always assume that the requests can be placed within a total order where the inter-arrival times are i.i.d.
We denote the request at a given time $t$ by  $R_t$, which consists of a pick-up location, a drop-off location, a pick-up time, and a drop-off time. 

In practice, it is common for paratransit services to use a pick-up window, e.g., they commit to picking passengers up at most 15 minutes before the requested pickup time and drop them off at most 15 minutes after the requested drop-off time.
%\Aron{this was very unclear to me (request ``relatively later'' could be interpreted as requesting late, i.e., just before taking the ride), suggest clarifying that we mean well in advance (next sentence clarifies, but reader may struggle with this one before the next one)}relatively later than other on-demand ride services (e.g., taxis). For example, customers can call in the morning to request for a ride during early afternoon. However, customers can also request for rides sooner, e.g., in an hour. We assume that the requests are placed one at a time; in principle, we can always \Aron{so, are we considering the discrete-time model? still not crystal clear at this point}discretize time fine enough such that this assumption approximately holds. We also assume that requests are drawn \Aron{acronym does not really fit the grammar of the sentence}i.i.d. from some known distribution $D$. We denote \Aron{``the'' (since there is only one)}a request made at time $t$ by $R_t$, which consists of a pick-up location, a drop-off location, a pick-up time, and a drop-off time. In practice, it is common for paratransit services to use a \Aron{pick-up vs. pick-up inconsistency (recurring)}pickup window, e.g., they commit to picking passengers up at most 15 minutes before the requested pickup time and drop them off at most 15 minutes after the requested drop-off time.
For a request $R_t$, we refer to such time points as the earliest pick-up time (denoted by~$e_t$) and the latest drop-off time (denoted by $p_t$). Naturally, the latest drop-off time must be greater than the sum of the earliest pick-up time and the minimum time it takes to travel to the  drop-off location from the pick-up location. We assume that requests follow such constraints. In practice, the application used for making requests or a human operator can enforce such constraints. Note that the time windows
specified by the customers are treated as strict guidelines making this a pickup and delivery problem with time-windows (PDPTW)~\cite{dumas1991pickup}.
%are specified in advance to the customers and are strict in our settings.
In case constraints cannot be met, the system \textit{rejects} requests. Given this setting, the goal of the decision-maker is to maximize the total number of requests that can be served in a day while ensuring that the pick-up and drop-off constraints are~met.

%To compute feasibility, we use a heuristic-based solution approach to the pickup and delivery problem with time-windows (PDPTW)~\cite{dumas1991pickup}.

% \Aron{this problem statement is not very clear (e.g., are requests assigned permanently to vehicles? does not seem to be specified; what are the travel times? do they vary? are they known? does not seem to be specified), reader may feel confused at this point}
% \Aron{also, this paragraph is quite long}
% Given the problem setting, the goal of the decision-maker is to maximize the total number of requests that can be served in a day while ensuring that the pick-up and drop-off constraints are~met. 

%\subsection{Decision Epoch}

\subsection{Route-based Markov Decision Process}

% An overview of the problem space is provided in figure~\ref{fig:overview2}. In this work we are focused on the online dispatch optimization problem which consists of servicing same-day requests in addition to the day-ahead scheduling problem which generates an initial set of route plans for the vehicles. There are many offline solvers for this problem and our approach to online dispatching is agnostic to the offline solver used.

\begin{figure*}[t]
\includegraphics[width=\textwidth]{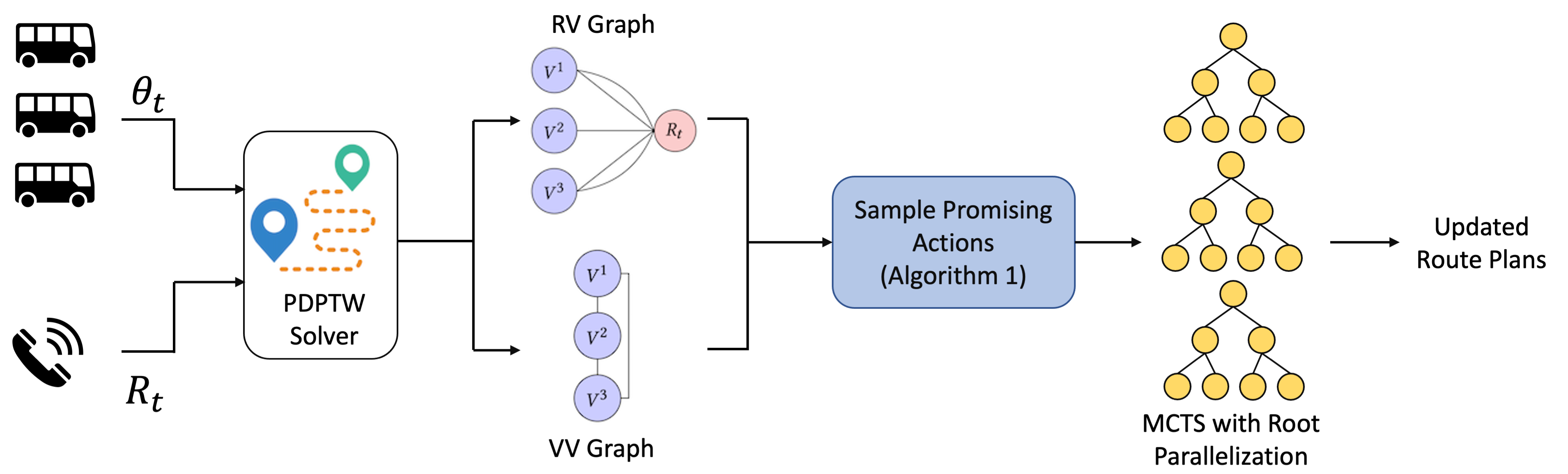}
\caption{An overview of MC-VRP. A decision epoch corresponds with a new trip request $R_t$ . We generate request-vehicle (RV) and vehicle-vehicle (VV) graphs by combining a heuristic based PDPTW solver with metrics to quickly estimate the utility of route plans. 
We then select promising actions from the graphs by sampling independent sets of high weights to be evaluated by an online approach based on MCTS.}
\label{fig:overview}
\end{figure*}

Our problem consists of a set of identical vehicles, denoted by $V$, each with a capacity of $c$ passengers. The set of all possible locations in the area is represented by the graph $G = (L, E)$, where $L$ denotes the set of vertices (or locations) in the graph, and $E$ denotes the set of edges weighted by travel time.
%(i.e., roadways between locations).
% \todo[inline]{Aron: this definition of route plans (and the corresponding notation) are not required to define the problem; in fact, the next paragraph, which defines the problem informally at the end, does not use the notion of route plan; further, it is difficult to define route plans without having first introduced the requests (route plan consists of pickup and dropoff locations, but those have not been introduced, so we use vague ``arbitrary vertices''); strongly recommend moving route plans later, either after introducing requests but before defining the problem informally at the end of this subsection, or into the next subsection (since the informal problem definition does not use the notion of route plans either)}
A route plan for vehicle $v^i \in V$ at time $t$ is denoted by $\theta^i_t$, which is an ordered sequence of pick-up/drop-off locations that the vehicle needs to visit in its route. Therefore, a route plan~$\theta^i_{t}$ can be represented as  an ordered set $\{l^1, l^2, \ldots\}$, where each $l^i \in L$ is a vertex of graph $G$. At any given time $t$, the set of all feasible route plans is denoted by $\Theta_t$ (we define feasibility below). We assume that vehicles start operations at a depot and return to the depot at the end of the day ($depot \in L$).

To solve the problem of identifying the best routes for all vehicles, we model the dynamic vehicle routing problem (DVRP) as a Markov decision process (MDP) based on prior work done by \citeauthor{ulmer2017route}~\cite{ulmer2017route}. An MDP is defined by a $4$-tuple $\{S,A,P,\gamma\}$, where $S$ is a set of states that capture relevant information for decision-making, $A$ is a set of control choices or actions, $P$ is a state-action transition function, and $\gamma$ is a reward function that defines the utility of taking an action at a given state~\cite{kochenderfer2015decision}. The route-based MDP formulation is beneficial in DVRP settings where actions involve assigning the current request to a vehicle and optimizing the entire route under consideration. Since our goal is to maximize the number of requests that can be served in expectation (with respect to the distribution $D$), optimizing the route as a whole towards this goal is a natural choice.

\textbf{Decision Epoch} We define decision epochs as in prior work by \citeauthor{joe2020deep}~\cite{joe2020deep}. A decision epoch occurs at the time a request is received. Between requests, the environment 
%\Aron{still not very clear what time is (referring to ``time steps'' suggests to the reader that time is discrete, but then the next sentence states that time is discrete, so reader will wonder why we have steps?)}every time step. In principle, the environment we consider 
evolves in continuous time, e.g., the vehicles move continuously, and requests can arrive at any point in time. 
% \ad{Note that in practice the system continues to evolve while a decision is being taken. Further, even though the requests are sparse a next request can arrive while the previous requests is being processed. We need to say something how do we handle this. Or perhaps we can say that we freeze state updates while we take an action. for example, if someone is being dropped off, we will synchronize and ensure while the decision is being made we can approximate that the person being dropped off is still in the vehicle }
%In practice, we can
%\Aron{model is continuous but practice is discretized? may be major confusion to the reader}discretize time finely enough based on the problem at hand. 
At each decision-epoch, the decision-maker takes an %\Aron{perhaps add a note stating that states and actions are defined below}
action (states and actions are defined below). The effect of the action results in a state transition, which consists of two parts --- a transition from a pre-decision state to a post-decision state, and from the post-decision state to the next pre-decision state~\cite{ulmer2017route,powell2007approximate}.

\textbf{State} We denote the set of states by $S$. We use $s_t$ to denote the pre-decision state at time $t$, which includes the vehicle locations, current route plans for all vehicles, and details about passengers aboard the vehicles. Note that while it suffices to keep track of the route plan for some formulations, we must track drop-off times for passengers already on board as actions may dynamically change routes. Formally, we represent the state $s_t$ by the tuple $(R_t, R, vloc_t, \theta_t)$, where $R_t$ is the new trip request at time epoch $t$, $R=\{R^1, \dots, R^{\mid V \mid}\}$ is the set of requests assigned to each of the vehicles, $vloc_t$ is a vector of location of all the vehicles at time $t$, and $\theta_t = \{\theta^1_t,\dots,\theta^{\mid V \mid}_t\}$ is the current route plan for each of the vehicles. 
% The request $R_t$ pickup location, a drop-off location, the earliest time that the passenger can be picked up (denoted by $e_t$, and the latest time that the passenger can be dropped off (denoted by $p_t$). 
The post-decision state is denoted by $s^t_x$, which denotes the effect of an action $x$ on $s_t$, and includes the updated route plan~\cite{ulmer2017route}.
% Note that the post-decision state, upon the effect of an action at decision-epoch $t$, consists of a route plan indexed by $t+1$, as the route plan includes planned actions beginning at the next epoch~\cite{ulmer2017route} \textcolor{red}{MW: I don't think the post decision route plan at time $t$ is the same as the pre-decision route plan at $t+1$ since vehicles traverse their plans from post-decision state to the next pre-decision state some of the nodes will be visited and dropped from the route plan during this transition}. 

We associate some additional information with each route plan. Consider a route plan $\theta^{i}_t$ for vehicle $v^i \in V$. For each location $l^j$ in the route plan, we associate four pieces of information. First, let $a(\theta^i_t, l_j)$ denote the planned arrival time of the vehicle $v_i$ at location $l^j$. We can calculate the arrival time from a pre-computed travel time matrix (or a router). Second, let $e(\theta^i_t, l^j)$ be the earliest time service may begin at this location. If $l^j$ is a pick-up location, $e(\theta^i_t, l^j)$ is equal to the earliest pick-up time for the customer; otherwise, we set it to some default value. Third, $p(\theta^i_t, l^j)$ denotes the latest time at which service is desired at this location. If $l^j$ is a dropoff location, $p(\theta^i_t, l^j)$ is set to the latest drop-off time for the customer associated with the location; otherwise, we set it to some default value. We also maintain the number of passengers on-board the vehicle at each location as $w(\theta^i_t, l^j)$. Lastly, let $y^j$ be a binary variable set to 0 if $l^j$ is a pickup location and set to 1 otherwise.  

% specifies whether this is a pickup or dropoff location. If its a pickup location then $p(\theta^m_{i,j})=0$. If its a dropoff location then $p(\theta^m_{i,j})=1$.

\textbf{Actions} We denote the set of all feasible actions at time $t$ by $X_t$. An arbitrary action in $X_t$ involves assigning the new trip request $R_t$ to a vehicle $v^i \in V$ and subsequently updating the route plan of \textit{all} the vehicles. Note that in our problem setting, $X_t$ simply reduces to $\Theta_t$, the set of all feasible route plans at time $t$. Updating the route plan can entail changing the order in which existing requests are picked up/dropped off in a vehicle's route plan or swapping requests that have not been picked up between vehicles. While allowing swapping between vehicles increases the complexity of the problem significantly, we include such actions nonetheless to maximize utility. The action space is, therefore, the set of \textit{all} feasible route plans. A feasible route must meet three requirements: first, the pick-up location for each trip requests must appear before the corresponding drop-off location in the route plan. Second, the pick-up time for a request must not occur before its earliest pick-up time, and finally, the drop-off time must not occur after its latest possible drop-off time.  

% We define these conditions for an action $R_t$ as:

% %To ensure that the pickup time does not occur before the earliest passenger pickup time ($e_n$) and the dropoff time does not occur after the latest possible dropoff time ($l_n$) we have the following two time constraints.

% \begin{equation}\label{eq: early window constraint}
%     e(\theta_{i,j}) \leq a(\theta_{i,j}) \text{ if } p(\theta_{i,j})=0
% \end{equation}

% \begin{equation}\label{eq: late window constraint}
%     a(\theta_{i,j}) \leq l(\theta_{i,j}) \text{ if } p(\theta_{i,j})=1
% \end{equation}

 \textbf{State Transitions and Rewards} At decision-epoch $t$, the decision-maker can take an action $x^j \in X_t$ (say), which results in a transition from the pre-decision state $s_t$ to the post-decision state $s_t^x$. Then, the system transitions from $s_t^x$ to the next pre-decision state $s_{t+1}$, within which a new request might arrive. We assume that requests are drawn from a distribution $D$.
% The realization of a trip request defines each time epoch. The time between consecutive requests varies and is determined by the distribution $D$. Upon arrival, the request is assigned to a vehicle, and the current route plans are updated. Vehicles continue to move along their routes between states.
We refrain from defining the exact state-action transition function since we use a simulator (or a generative model) for online planning. The reward for an action $x^j \in X_t$, denoted by $\gamma(x^j)$, is simply the number of additional requests served by the action. Since only one request arrives at any decision epoch, the reward simplifies to 1 if the request can be accommodated and 0 otherwise. Since paratransit services require strict adherence to time windows, note that some requests can be rejected. In practice, human operators can suggest alternative choices to customers in such situations. %We present a notation lookup table in the technical appendix (section~\ref{sec_appendix:notation}) for convenient reference to notation. 

\section{Approach}

We provide an overview of MC-VRP in Figure~\ref{fig:overview}. Our approach consists of the following three broad components: \textbf{1)} For a given state of the MDP, we sample feasible and promising actions by exploiting the structure of the problem. To compute feasibility, we use a heuristic-based solution approach to the PDPTW. To compute the potential utility of a feasible action, we introduce two heuristics, one based on passenger travel times and another based on \textit{budget} (or slack) in route plans. \textbf{2)} We compute weighted graphs based on the feasible actions and their potential utilities. Specifically, we create vehicle-vehicle (VV) and request-vehicle (RV) graphs (we describe the graphs below). Then, we generate promising actions by sampling independent sets from the graphs (based on the weights of the sets). \textbf{3)} The sampled actions are then used by our online non-myopic planner based on Monte Carlo tree search. To build the search tree into the future, we sample future requests from a data-driven generative model. Finally, to lower computation time, we utilize pre-computed samples of requests and \textit{root parallelization}~\cite{mukhopadhyay2019online} to efficiently explore the search space and recommend an action for the given state. We describe each component of our approach in detail below.

%\begin{figure*}[t]
%\includegraphics[width=0.9\textwidth]{images/MC-VRP_updated_3.png}
%\caption{An overview of MC-VRP. A decision epoch corresponds with a new trip request $R_t$ . We generate request-request (RV) and vehicle-vehicle (VV) graphs by combining a heuristic based PDPTW solver with metrics to quickly estimate the utility of route plans. 
%We then select promising actions from the graphs by sampling independent sets of high weights to be evaluated by an online approach based on MCTS.}
%\label{fig:overview}
%\end{figure*}

\subsection{PDPTW Solver}\label{section:tspSolver}

At each decision epoch, our goal is to optimize existing vehicles' routes to accommodate a new request. First, we check whether a vehicle can accommodate the request given its current route plan. Recall that accommodating a request in our setting requires strict adherence to time windows. We use a heuristic-based solver for the pickup and delivery problem with time-windows (PDPTW). The solver enables us to check if the current request can be accommodated in a feasible route plan. The PDPTW is NP-hard~\cite{dumas1991pickup,furtado2017pickup}; as a result, we use a heuristic subroutine to solve it. Using a heuristic approach is critical in our setting; as we show below, the PDPTW solver needs to be invoked for the given state of the MDP as well as for states we sample as we look into the future. While any heuristic designed for solving PDPTW can be incorporated in our framework, we use the \textit{insertion heuristic} \cite{alonso2017demand}, which seeks to insert the pickup and dropoff locations of the new request within the existing route plan. We introduce some additional notation to describe the PDPTW solver. Consider a vehicle $v^i \in V$ that has an assigned route $\theta^i_t$ at time $t$. For a new request $R_t$, we use $PDPTW.feasiblePlans(\theta^i_t, R_t)$ to denote a module that returns the set of all feasible route plans for the vehicle $v^i$ that include the new request. Also, let $PDPTW.bestFeasiblePlan(\theta^i_t, R_t) = \text{argmax}_{\theta} U_{\omega}(PDPTW.feasiblePlans(\theta^i_t, R_t))$, which denotes a function that computes the utility of each feasible route plan generated by the PDPTW solver based on a specific utility function $U$ and a metric $\omega$, and returns the one with the highest utility. For example, a metric could be the total travel time of the route, and the utility function, in the simplest case, can be the identity function.

\subsection{Handling Exponential Action Space}\label{subsection: handling exponential action space}

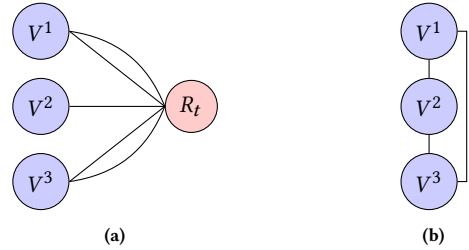
\begin{figure}[t]
    \centering
    %\resizebox{.85\linewidth}{!}{%
    \begin{subfigure}[b]{.5\linewidth}
    \centering
    \begin{tikzpicture}
        \tikzstyle{vnode}=[draw=black,fill=blue!20,shape=circle,minimum size=.5cm]
        \tikzstyle{rnode}=[draw=black,fill=red!20,shape=circle,minimum size=.5cm]
    
        % nodes
        \node[rnode](r1) at (1,-1) {$R_t$};
        \node[vnode](v1) at (-1,0) {$V^1$};
        \node[vnode](v2) at (-1,-1) {$V^2$};
        \node[vnode](v3) at (-1,-2) {$V^3$};
    
        % edges
        \draw (r1.west) to[bend right] (v1.east);
        \draw (r1.west) edge[-] (v1.east);
        \draw (r1.west) edge[-] (v2.east);
        \draw (r1.west) edge[-] (v3.east);
        \draw (r1.west) to[bend left] (v3.east);
    \end{tikzpicture}
    \caption{}
    \end{subfigure}%
    \begin{subfigure}[b]{.5\linewidth}
    \centering
    \begin{tikzpicture}
        \tikzstyle{vnode}=[draw=black,fill=blue!20,shape=circle,minimum size=.5cm]
        \tikzstyle{rnode}=[draw=black,fill=red!20,shape=circle,minimum size=.5cm]
    
        % nodes
        \node[vnode](v1) at (0,0) {$V^1$};
        \node[vnode](v2) at (0,-1) {$V^2$};
        \node[vnode](v3) at (0,-2) {$V^3$};
    
        % edges
        \draw (v1.south) edge[-] (v2.north);
        \draw (v2.south) edge[-] (v3.north);
        \draw (v1.east) -- (.5, 0) -- (.5, -2) -- (v3.east);
        %\draw (v1.north) -| (v3.north);
    \end{tikzpicture}
    \caption{}
    %}%
    \end{subfigure}
\caption{(a) RV-graph ($G_{RV}$): there is an edge between a request and a vehicle for every feasible route plan in which a vehicle can service the new request. (b) VV-graph ($G_{VV}$): an edge between vehicle $v^i$ and $v^j$ represents the swap with the highest utility between the two vehicles.}
\label{fig: RVV graph}
\end{figure}

%\begin{figure}[t]
%\centering
%\begin{subfigure}{0.3\columnwidth}
%    \includegraphics[height=1in]{images/feasibility2.png}
%    \caption{}
%    \label{subfig: RV graph}
%\end{subfigure}
%\begin{subfigure}{0.3\columnwidth}
%    \includegraphics[height=1in]{images/feasibility1.png}
%    \caption{}
%    \label{subfig: VV graph}
%\end{subfigure}
%\caption{(a) RV-graph ($G_{RV}$): there is an edge between a request and a vehicle for every feasible route plan in which a vehicle can service the new request. (b) VV-graph ($G_{VV}$): an edge between vehicle $v^i$ and $v^j$ represents the swap with the highest utility between the two vehicles.}
%\label{fig: RVV graph}
%\end{figure}

A feasible action in our problem corresponds to a set of route plans for all vehicles, given that one of them can accommodate the new request in consideration. In case no feasible action is found, the request is rejected. Additionally, we design our action space to let vehicles swap requests from their assigned routes that have not been picked up (we describe how we use the PDPTW solver to this end below). As a result, the number of possible actions for a given state is combinatorially large; on average, an arbitrary state in our MDP has $10^{22}$ possible actions.  Such an action space is infeasible to explore in an online setting. To address this challenge, we introduce an approach that enables us to sample promising actions from the set of feasible actions. We start by introducing two heuristic metrics that can be used to gauge the long-term utility of a route plan quickly.

% The number of possible route plans is therefore exponential and is infeasible to evaluate efficiently with local search heuristics alone in an online setting. In this section we provide a methodology for generating a subset of promising feasible actions to be evaluated further through MCTS. We leverage the structure of the problem to filter out actions which are not likely to be good choices and therefore narrow the search space for our non-myopic evaluation. In this way, the MCTS search can better focus on high quality solutions. This pruning increases performance and efficiency in the online approach.

\noindent
\textbf{1.) Maximizing the budget to serve future requests:} Our goal is to maximize the number of requests the vehicles serve on a given day while following the specified time constraints. Intuitively, a vehicle can accommodate future requests in an existing route plan if there is sufficient \textit{room} (time) in the route. To capture this idea formally, we build upon prior work by \citeauthor{ulmer2018budgeting}~\cite{ulmer2018budgeting} to extend the idea of a budget-based heuristic to DVRPs with capacity and time window constraints. Our budget-based heuristic captures the idea that maximizing the time a vehicle has no passengers on board also maximizes the slack to serve future requests. We define the budget-based utility for a route plan for vehicle $v^i \in V$ at time $t$ as
\small
\begin{equation}\label{equation:budget utility}
    \overline{b}(\theta^i_t)=t_{max }- t- \sum_{j\in\{1,\dots,\mid\theta^i_t\mid -1\}} \mathbbm{1}(w(\theta_t^i,l^j)>0)\big\{a(\theta_{t}^i,l^{j+1}) - a(\theta_{t}^i,l_{j})\big\}
\end{equation}
\normalsize
where $t_{max}$ denotes the maximum time up to which the vehicle is available to serve requests (e.g., end of a day), $(w(\theta_t^i,l_{j}))$ denotes the number of passengers on board vehicle $v^i$ from location $l^j$ in its route plan to the next location, $\mathbbm{1}()$ denotes the indicator function, and $a(\theta, l^k)$ denotes the time a vehicle operating under a route plan $\theta$ reaches location $l^k$. The summation in the equation~\ref{equation:budget utility} represents the total time in the route plan for which at least one passenger is on-board.

\noindent
\textbf{2.) Minimizing passenger travel time:} An alternative to the budget-based heuristic is to minimize passenger travel time (PTT). Intuitively, by minimizing passenger travel time, we maximize the available capacity in each vehicle over the time horizon. We define the utility of the PTT-based heuristic for a route plan $\theta_t^i$ as $\overline{PTT}(\theta_t^i)$ for vehicle $v^i \in V$ at time $t$ as shown in equation:
\small
\begin{equation}\label{equation:ptt utility}
    \overline{PTT}(\theta^i_t)=\sum_{j\in\{1,\dots,\mid\theta^i_t\mid -1\}} w(\theta_t^i,l_{j})*(a(\theta_{t}^i,l_{j+1}) - a(\theta_{t}^i,l_{j}))
\end{equation}
\normalsize

In this case $\overline{PTT}(\theta^i_t)$ is the summation of the number of passengers on board after picking up the passenger at location $j$, represented by $w(\theta_t^i, l_j)$, multiplied by the time to reach the next location, $j+1$. Therefore, by minimizing $\overline{PTT}(\theta^i_t)$ we maximize the number of seats available to incorporate future requests. 

Having described metrics to assess the potential utility of a specific vehicle route for a given vehicle, we now introduce an approach to sample feasible route plans. We begin by describing two graphs we construct based on the new request that arrives at a decision epoch and the existing route plans of the vehicles.

\textbf{RV graph:} At each decision epoch, we first generate a graph that incorporates which vehicles can service the new request. Our idea is based on prior work by \citeauthor{alonso2017demand}~\cite{alonso2017demand}. We denote the RV graph by $G_{RV}=(L_{RV}, E_{RV})$, where $L_{RV}$ denotes a set of vertices and $E_{RV}$ denotes a set of edges. To create the RV graph, we add a node for each vehicle $v_i \in V$. We add an additional node to denote the request $R_t$. Then, for each feasible vehicle route that can accommodate the request, we add an edge between the node denoting the request and each node representing a vehicle. We denote a specific edge in $E_{RV}$ by $\edgerv(i,j)$, which denotes the $j$th feasible route plan for vehicle $v^i$ that can accommodate the request under consideration. The feasible route plans are generated by the module $PDPTW.feasiblePlans(\theta^i, R_t)$ (for vehicle $v^i \in V$ and request $R_t$). We associate two pieces of information with each edge. First, for an edge $\edgerv(i,j)$, we use $U_\omega (\edgerv(i,j))$ to denote the utility of the edge based on a utility function $U$ and metric $\omega$. Also, we use $\theta(\edgerv(i,j))$ to denote the updated route plan corresponding to the edge $\edgerv(i,j)$. We show an example of an RV graph in figure~\ref{fig: RVV graph}.

\textbf{VV-graph:} While an RV graph is useful to represent which vehicles can accommodate the new request in their existing route plans, it is possible that vehicles might want to swap requests that have not been picked up to maximize utility. Note than in a non-myopic setting, the route plans are optimized at each decision-epoch to maximize the expected utility with respect to $D$ (the request arrival distribution). However, a specific realization of $R_t$ presents an opportunity for vehicles to re-plan and potentially swap requests. In order to represent such swapping actions, we create an undirected vehicle-vehicle (VV) graph at each decision epoch. We denote a VV graph by $G_{VV}=(L_{VV}, E_{VV})$. 

To construct the graph, we add a node for each vehicle $v^i \in V$. Edges between two vehicles denote potential swaps of requests that have not been picked up. An edge $\edgevv(i,j,k) \in E_{VV}$ denotes the potential swap of request $R_k$ from vehicle $v^i$ to vehicle $v^j$. Note that a swap action creates two new route plans, one for each vehicle. We use $\theta(\edgevv(i,j,k), v^i)$ to denote the route plan for vehicle $v^i$ \textit{after} the swap (similarly, we use $\theta(\edgevv(i,j,k), v^j)$ to denote the route plan for vehicle $v^j$). The utility of a swapping action, denoted by $U_\omega (\edgevv(i,j,k))$ is denoted as the difference in utility of the updated route plans and the original route plans, i.e., the plans without the swapping action. As before, $U_\omega$ denotes utility computed according to a function $U$ and metric $\omega$. For example, using an identity utility function and a metric based on the budget heuristic introduced in equation~\ref{equation:budget utility}, $U_{budget}(\edgevv(i,j,k))=\overline{b}(\theta(\edgevv(i,j,k), v^j))+\overline{b}(\theta(\edgevv(i,j,k), v^i))-\overline{b}(\theta^{i}_t)-\overline{b}(\theta^{j}_t)$. Note that once a request is swapped, its pickup and dropoff can be inserted at multiple places within the existing route of the vehicle that receives it. For computational tractability, we choose the best insertion point using the module $PDPTW.bestFeasiblePlan$ (introduced in section~\ref{section:tspSolver}). We show an example of an VV graph in figure~\ref{fig: RVV graph}.

\textbf{Generating Feasible Actions:} A feasible action is an updated set of route plans for all vehicles that do not violate the time window or capacity constraints. We seek to sample a feasible route plan from the constructed graph. Our approach is motivated by prior work by \citeauthor{zalesak2021TR}~\cite{zalesak2021TR}.  Each edge in the RV graph represents an operation that creates an updated route plan for a vehicle that includes the new request. Additionally, each edge in the VV-graph represents a swapping operation which creates a new route plan for the two vehicles involved in the swap. Therefore, any selected edge in $G_{RV} \cup G_{VV}$ represents a feasible action. Additionally, multiple edges can be selected to generate a new action, \textit{if and only if} the set of selected edges includes only one of the edges from the RV graph. The rationale for such a condition is straightforward; a new request can only be assigned to one vehicle, and consequently, only one edge from the RV graph can be selected for a particular feasible action at each decision epoch. To generate potential feasible actions quickly, we sample independent sets (a set of edges that have no vertices in common) from $G_{RV} \cup G_{VV}$ that includes one edge from $G_{RV}$. Such an independent set guarantees feasibility, as we show below:

\begin{theorem}\label{theorem:ind}
Consider graphs $G_{RV}$ and $G_{VV}$ generated at decision epoch $t$. An independent set of edges from $G_{RV} \cup G_{VV}$ that includes one and only one edge from $G_{RV}$ must be a feasible action for the current state $s_t$ of the MDP. 
\end{theorem}

\begin{proof}
We first list the conditions for feasibility: 1) An action is feasible if it services the current request with adherence to time constraints, and 2) time constraints for existing requests are met (including potential swaps). An independent set with one edge from $G_{RV}$ meets condition 1 by construction---all edges from $G_{RV}$ service the current request and meet time constraints (recall that edges are checked for feasibility through the PDPTW solver). As the vehicle that services the request cannot swap requests (by the property of independence), all swaps sampled from $G_{VV}$ are also feasible---all edges in $G_{VV}$ are checked for feasibility through the PDPTW solver. Hence the set of independent edges must correspond to a feasible action for the given state.
\end{proof}

Notice that the only condition in which a chosen set of edges from $G_{RV} \cup G_{VV}$ might result in an infeasible action is the following: consider a situation in which the vehicle (say $v^i$) that is assigned the request (denoted by an edge from the RV graph) also engages in swapping and receives an additional request from a different vehicle (say $v^j$). While both the actions, namely the swapping action and the servicing of the new request, are checked for feasibility in isolation, vehicle $v^i$ might violate time constraints if it seeks to service both. In theory, it is possible to enumerate over each possible set of edges in  $G_{RV} \cup G_{VV}$ and check for feasibility; however, enumerating over all such sets is intractable in practice. Therefore, we point out that the vehicle that services the request \textit{can} engage in swapping and still maintain feasibility; however, ensuring that such a vehicle does not participate in swapping \textit{guarantees} feasibility. As a result, we use the heuristic of sampling independent sets (with one edge from the RV graph) that guarantee feasibility by construction. %We present a more detailed description of why independent sets ensure feasibility in the technical appendix (section~\ref{sec_appendix:ind_sets}.

Recall that our goal is to sample promising actions from the set of feasible actions since evaluating all feasible actions in an online setting is not tractable. We present our approach to selecting promising actions in algorithm~\ref{algo: feasible actions}. Our approach is based on sampling independent sets based on the cumulative sum of edge weights of the sets. First, we initialize an empty set of feasible actions $\Theta_t$ and an empty set of utilities $\mathbb{U}$. Then, we begin selecting an independent set by selecting an edge from $G_{RV}$ (step 3). Next, we initialize the utility for the route plan (that the independent set corresponds to) with the utility of the chosen edge (step 7). Then, we drop the corresponding vehicle node from $G_{VV}$ and consider swaps iteratively (step 9-10). The utility of the resulting action is calculated by adding the utility of serving the request and the swapping action (step 15). Finally, we return a subset of $K_{max}$ actions with the highest total utility (step 21-22) to be evaluated by the tree search, where $K_{max}$ is an exogenous parameter. 

\small
\begin{algorithm}[t]
\caption{Generating feasible actions}\label{algo: feasible actions}
\begin{algorithmic}[1]
\Require $\theta_t, R_t, K_{max}, M, G_{VV}, G_{RV}$
\State $\Theta_t \gets \{\}$ \Comment{empty set of feasible actions}
%\State $X_t \gets \{\}$ \Comment{empty set of feasible actions}
\State $\mathbb{U} \gets \{\}$ \Comment{empty set of utilities for each action}
\For{$\edgerv(i, j) \in E_{RV}$}
    \State $\theta \gets \theta_t$
    \State $\theta[i] = \theta(\edgerv(i,j))$
    \State $\Theta_t.append(\theta)$ 
    \State $U_x \gets U_{\omega}(\edgerv(i,j))$
    \State $\mathbb{U}.append(U_x)$
    \State $G_{VV}'=G_{VV}[\{v_k \in L_{VV} : k \neq i\}]$ 
    \While{$|E_{VV}'| > 0$}
        \State $\edgevv(m,n,k) = \argmax_{(m,n,k)}(U_\omega(\edgevv(m,n,k))$
        \State $\theta[m] = \theta(\edgevv(m,n,k), v^m)$ 
        \State $\theta[n] = \theta(\edgevv(m,n,k), v^n)$ 
        \State $\Theta_t.append(\theta)$ 
        \State $U_x = U_x + U_\omega(\edgevv(m,n,k))$
        \State $\mathbb{U}.append(U_x)$
        \State $G_{VV}'=G_{VV}'[\{v_k \in L_{VV} : k \neq m \And k \neq n\}]$ 
    \EndWhile
\EndFor
\State // Filter: $X_t$ is the $K_{max}$ actions with highest utility
\State $X_t=\Theta_t[argSort(U)[1:K_{max}]]$
\State \textbf{Return:} $X_t$
\end{algorithmic}
%\vspace{-0.3in}
\end{algorithm}
\normalsize

\subsection{MCTS Evaluation}\label{subsec: mcts evaluation}

% declaration of the new block
\algblock{ParFor}{EndParFor}
% customising the new block
\algnewcommand\algorithmicparfor{\textbf{for}}
\algnewcommand\algorithmicpardo{\textbf{do parallel}}
\algnewcommand\algorithmicendparfor{\textbf{end\ for}}
\algrenewtext{ParFor}[1]{\algorithmicparfor\ #1\ \algorithmicpardo}
\algrenewtext{EndParFor}{\algorithmicendparfor}

% The spatio-temporal nature and inherent uncertainty of online paratransit requests makes it challenging to evaluate potential actions in the context of future demand. Also, changes in congestion and road closures can introduce additional uncertainty into the planning process. When these underlying environmental dynamics change, the solution must take these updates into account to provide optimal recommendations.

% Most non-myopic approaches rely on hybrid offline-online solutions in which an offline component is trained on historical data and embedded in an online search~\cite{joe2020deep, ulmer2019offline}. These approaches' offline components typically require long training periods and must be re-trained each time the environment changes, making them unsuitable for highly dynamic environments. This motivates using MCTS, an online probabilistic search algorithm, to evaluate the long-term utility of potential actions. MCTS is an anytime algorithm, and any changing environmental conditions that are detected can immediately be incorporated into its underlying generative models for making decisions. 

Non-myopic approaches to DVRP rely on hybrid offline-online solutions in which an offline component is trained on historical data and embedded in an online search~\cite{joe2020deep, ulmer2019offline}. Offline components typically require long training periods and must be re-trained each time the environment changes, making them unsuitable for highly dynamic environments. This motivates us to use MCTS,  an online probabilistic search algorithm, to evaluate the long-term utility of potential actions. MCTS is an anytime algorithm, and any changing environmental conditions that are detected can immediately be incorporated into its underlying generative models for making decisions~\cite{pettet2021hierarchical}. 

MCTS represents the planning problem as a ``game tree,'' where states are represented by nodes in the tree. The current state is treated as the root node, and actions represent edges that mark transitions from one state to another. The fundamental idea behind MCTS is that the search tree can be explored asymmetrically, biasing the search toward actions that appear promising. To estimate the value of an action, MCTS simulates a ``rollout'' to the end of the planning horizon using a \textit{default policy}. In practice, the rollout policy only needs to be computationally cheap; a common method involves selecting actions randomly during rollout. As the tree is explored and nodes are revisited, each node's utility is estimated. As the search progresses, the estimates converge towards the true value of the node. This asymmetric tree exploration allows MCTS to search very large action spaces quickly. MCTS typically requires a few domain specific components: a generative model of the environment, the \textit{tree policy} used to navigate the search tree, and the \textit{default policy} used to estimate the value of a node. We describe each component below.
 
\textbf{Generative demand model:} The generative demand model (denoted by $E$) provides a method for sampling new requests as the tree is built into the future. We use a hierarchical modeling approach to sample trip requests based on historical data. We optimize model parameters based on maximum likelihood estimation. 
%For the sake of brevity, we present a detailed description of the generative model in the appendix (section~\ref{sec_appendix:generative_model}). 
We model two processes as part of the generative demand model. First, we model the distribution of the number of requests per day as a Gaussian distribution. We learn the parameters of the distribution by maximizing the likelihood of historical paratransit data. Second, to model individual trip requests, we aggregate historical trip requests and weigh each trip request by the number of times it is observed (often, some passengers in paratransit services request trips that have the same source, destination, and time every week). To sample a sequence of trip requests for a day, we perform the following steps: 1) we sample the number of requests (say $m$) from the learned Gaussian distribution over trip requests. 2) We sample $m$ requests from the weighted aggregation of trip requests.% The second distribution is the aggregation of all the trips, where each unique trip is weighted by the number of occurrences of that trip since a particular customer often requests the same trip on separate days of the week. A single chain is generated by first sampling the number of trips in a day from $g_{day}(len(\mathcal{R})|u_{day}, \sigma_{day})$, and then $len(\mathcal{R})$ trips are sampled from the weighted aggregation of all trips. 
This process is repeated a number of times to generate multiple sampled \textit{chains} offline. During inference at decision epoch $t$, we provide a method $E.sample(t, n)$ which samples $n$ chains uniformly at random from $E$ and returns the requests in each chain that occur after the current time $t$. 

\textbf{Search policy:} We use the standard Upper Confidence bound for Trees (UCT)~\cite{kocsis2006bandit} to navigate the search tree and decide which nodes to expand. When expanding a node, we use algorithm~\ref{algo: feasible actions} to sample feasible actions for the given state. When working outside the MCTS tree to estimate the value of an action during rollout, we rely on a default policy. This is a lightweight policy which is simulated up to a time horizon and the utility of the simulation is propagated up the tree. Our default policy is a greedy assignment---for a given state, we choose the edge with the highest myopic utility from the RV-graph. To ensure tractability, we do not incorporate swapping requests between vehicles into our rollout policy; this saves time during MCTS evaluation as the VV graph is not generated during rollout.

\textbf{Root parallelization:} Given that the sampled paratransit requests can be both sparse and highly uncertain in time and space, sampling one chain of requests might not adequately represent future demand. To handle this uncertainty, we use \textit{root parallelization}, which involves sampling many chains, and instantiating a separate MCTS tree for each with the current request as the root node. Crucially, each tree is explored in parallel. After execution, the score for each of the actions from the common root node is averaged across trees. Then, the action with the highest average score across the trees is returned as the selected action.

\small
\begin{algorithm}[t]
\caption{MCTS evaluation}\label{algo: mc-eval}
\begin{algorithmic}[1]
\Require $X_t, S_t, E, n_{chains}$
\State $eventChains = E.sample(S_t, n_{chains})$ \Comment{sample chains}\label{algo:mcts_sample}
\State $A = MCTS(X_t, eventChains)$ \Comment{action scores}\label{algo:mcts_mcts}
\State $\bar{A} = \{\}$
\ParFor{$a \in A$}
    \State $\bar{A}.append(mean(a))$ \Comment{aggregate across chains}\label{algo:mcts_agg}
\EndParFor
\State // Return action with highest action score
\State \textbf{Return:} $argmax_{x_i \in X}(\bar{A}[i])$ \label{algo:mcts_return}
\end{algorithmic}
\end{algorithm}
\normalsize

The process for evaluating and selecting an action is provided in algorithm~\ref{algo: mc-eval}. The algorithm takes $X_t$ from the feasible actions component as well as the current state $S_t$, the generative model $E$ and the number of chains to sample $n_{chains}$. First, the set of $eventChains$ is sampled from the generative model (line \ref{algo:mcts_sample}). Second, a tree is instantiated and MCTS is performed in parallel on each of the trees. This processed is represented by $A = MCTS(X_t, eventChains)$, where $A$ is a two dimensional tensor of size $|X_t| \times n_{chains}$ (line \ref{algo:mcts_mcts}). In this sense, each row in $A$ represents a feasible action and the columns represent the chains. The mean of each row in $A$ represents the action score and is stored in $\bar{A}$ (line \ref{algo:mcts_agg}). Finally, the action with the maximum action score in $\bar{A}$ is returned (line \ref{algo:mcts_return}).

%\begin{algorithm}[t]
%\caption{MC-VRP}\label{algo: mc-vrp}
%\begin{algorithmic}[1]
%\Require $R_t, S_t, \theta_t, E, K_{max}, n_{chains}, G_{RV}, G_{VV}, swap$
%\State // $G_{VV}$ only required if $swap = true$
%\If{$swap$}
%    \State $X_t=FASwap(\theta_t, R_t, K_{max}, M, G_{RV}, G_{VV})$
%\Else
%    \State $X_t=FA(\theta_t, R_t, K_{max}, M, G_{RV})$
%\EndIf
%\State $x_t=MCEval(X_t, S_t, E, n_{chains})$
%\State \textbf{Return:} $a$
%\end{algorithmic}
%\end{algorithm}

%The full decision making process for MC-VRP is as follows. MC-VRP requires the request ($R_t$), the current state ($S_t$), route plans ($\theta_t$), the generative model $E$, $K_{max}$, $n_{chains}$, the RV-graph ($G_{RV}$), the VV-graph ($G_{VV}$) and a boolean parameter $swap$ which represents whether swapping will be considered. Note that $G_{VV}$ is only required if $swap=true$. If we are considering swapping, then $X_t$ is generated from FA-Swap. Otherwise, $X_t$ is generated from FA. Lastly, the action returned from MC-Eval is selected.

%First, we generate the RV-graph ($G_{RV}$) as shown in figure~\ref{subfig: RV graph}. If swapping is allowed, we then generate the VV-graph ($G_{VV}$) as shown in figure~\ref{subfig: VV graph}.
\section{Experiments and Results}

\definecolor{drlsa}{rgb}{0.0, 0.6, 0.0}
\definecolor{pnas}{rgb}{0.0, 0.0, 0.8}
\definecolor{greedyptt}{rgb}{0.43, 0.21, 0.1}
\definecolor{greedybudget}{rgb}{0.0, 0.3, 0.4}
\definecolor{mcvrpptt}{rgb}{0.3, 0.4, 0.8}
\definecolor{mcvrpbudget}{rgb}{1.0, 0.0, 0.0}

% We evaluate MC-DVRP using real-world paratransit data against three baseline approaches. We first describe the data, experimental setup, and baseline approaches and then present experimental results. 
%\subsection{Experimental Setup and Data Description}\label{subsection: experimental setup and data description}

In this section we describe the data, experimental setup, system parameters, baselines and results.

\subsection{Data Description and Pre-Processing}\label{sec_appendix:pre_processing}

\begin{figure}[t]
    \centering
    \includegraphics[width=\columnwidth]{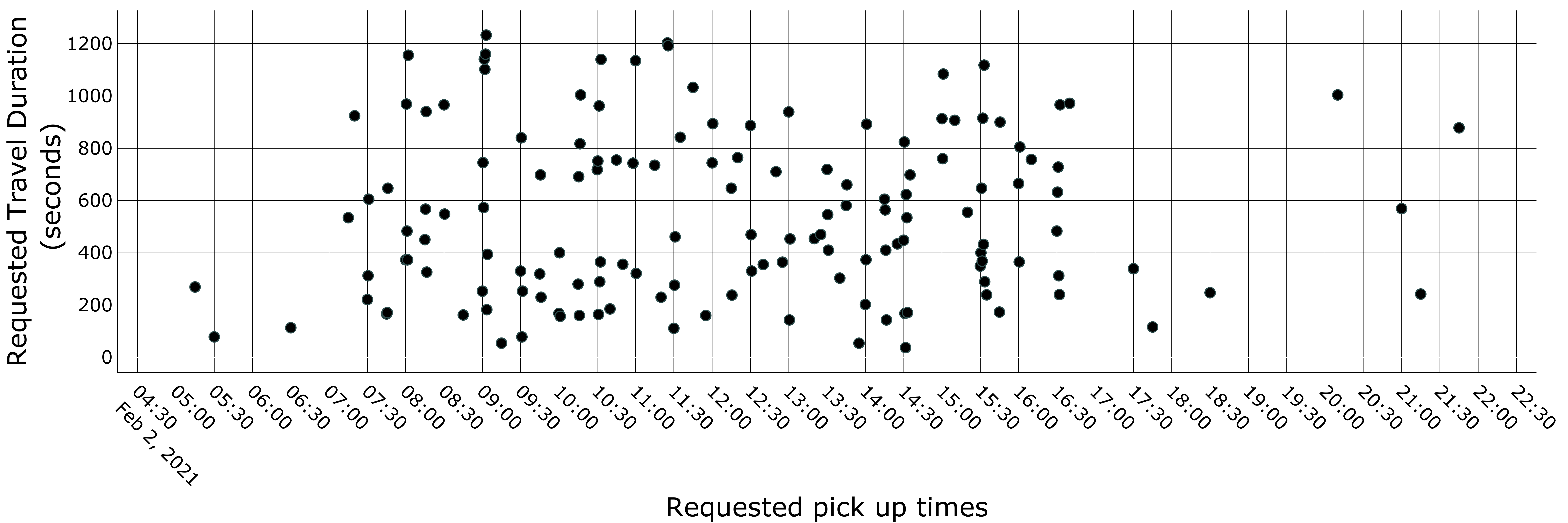}
    \caption{The temporal distribution of requests and the requested travel duration for the requests from one day.}
    \label{requestset}
\end{figure}

\textbf{Paratransit Data:} We acquired six months of paratransit trip requests between January 1, 2021 and July 1, 2021 from CARTA. Our dataset consists of a total of 25,843 trip requests. Each request in our dataset consists of a geo-coordinate (longitude, latitude) for the pickup location and dropoff location, and also the requested pickup time. As an example, we show the temporal distribution of one of the days and the requested travel duration for each request in figure~\ref{requestset}.
For pre-processing, each pickup and dropoff location was assigned to the nearest node in the road network graph. We used the haversine distance to calculate the distance between two geo-coordinates. We filtered out any requests that had a pickup or dropoff location that was farther than 200 meters from the nearest node in the road network; this process filtered out approximately 5\% of the requests. Of the 25843 trips in the original dataset, 24543 remained after the distance filter. As paratransit services are significantly more sparse on weekends, we included only weekday trips for our analysis that spanned 129 days (94\% of the trips in our data occur on weekdays).% Second, we only included weekday trips. There were 23052 trips in the final dataset spanning 129 days.
We randomly selected 15 days from the dataset for the evaluation (test) set. The remaining 114 days were use to compute 100 synthetic days worth of trip requests (i.e., chains) for the generative model using the procedure outlined in section~\ref{subsec: mcts evaluation}. Three more chains were generated as a calibration set for parameter tuning.

\textbf{Road Network and Travel Time Matrix:} We use OpenStreetMap (OSM)~\cite{OpenStreetMap} for the road network for the area under consideration and
OSMNX~\cite{boeing2017osmnx} to generate a routing graph of the road network with travel time for edge weights.
Unless otherwise noted, the experiments used the free flow speed as edge weight. Then, we calculated the shortest paths between all pairs of network to generate a matrix of travel times. The travel time matrix is generated offline and therefore, provides constant lookup time for querying travel times between arbitrary locations in the area under consideration. Additionally, we collected historical traffic data from INRIX~\cite{inrix} to estimate typical travel times during times of high congestion. We use congestion data for evaluating the robustness of the proposed approach.

%\textbf{Paratransit dataset:} We acquired six months of paratransit trip requests between January 1, 2021 and July 1, 2021 from CARTA. Our dataset consists of a total of 25,843 trip requests. Each request in our dataset consists of a geo-coordinate (longitude, latitude) for the pickup location and dropoff location, and also the requested pickup time.

%Since the dataset did not consist of requested dropoff time, we set the requested dropoff time to sum of the requested pickup time and the shortest time duration to travel from the pickup to dropoff location. 

% \textcolor{red}{MW: we should highlight the dataset more since we are going to make an anonymized version available. Include a figure on distribution of requests and request duration over time.} 
\textbf{Road network and travel time matrix:} We use OpenStreetMap (OSM)~\cite{OpenStreetMap} for the road network for the area under consideration and
OSMNX~\cite{boeing2017osmnx} to generate a routing graph of the road network with travel time for edge weights. To evaluate robustness we use INRIX traffic data to generate new travel time matrices that represent typical congested patterns in the city. The INRIX dataset included average roadway speeds per hour for each day in the week and included an OSM ID which was mapped to the OSM road network. 
%We describe our approach for generating the travel time matrix in the technical appendix (section~\ref{sec_appendix:pre_processing}).
% The INRIX data is used in the robustness experiments to generate new travel time matrices that represent typical congested patterns in the city.

%\textbf{Data pre-processing:} We processed the paratransit data and mapped each request to the road network graph. We describe pre-processing steps in the technical appendix (section~\ref{sec_appendix:pre_processing}). 
%The processed data spanned 129 days. We randomly selected 15 days from the dataset for the evaluation (test) set. The data from the rest of the days was used for training the generative model and hyperparameter tuning.

\subsection{System Parameters}

Our experimental parameters are as follows: we vary the number of vehicles from 3 to 5 as we find that 5 vehicles can serve 100\% of the trip requests in our setting using MC-VRP. The vehicle capacity is set to 8, which is in accordance with the capacity of typical paratransit vans. The request arrival time, defined as the time before the requested pickup time that the request is available to the system, is set to 60 minutes. In practice, such requests can also be made before 60 minutes. The time window, i.e., the amount of time before the requested pickup time that a request can be picked up is set to 15 minutes (in accordance with settings used by CARTA). Additionally, when a customer requests a trip we provide an estimated dropoff time which is the sum of the requested pickup time and the minimum travel time to the dropoff location. The late time window is 15 minutes after the estimated dropoff time which is again, in accordance with settings used by our partner agency. Additionally, we reiterate that as the time windows are hard constraints, a trip is only feasible if the passenger is picked up and dropped off between the early pickup window and late dropoff window.

%Additionally, the late time window is the requested pickup time plus the shortest path to the dropoff location plus 15 minutes. In this way, the early pickup window is 15 minutes before the requested pickup and the late dropoff window is 15 minutes after the requested dropoff time (i.e. requested pickup time plus shortest distance to dropoff location). 

Since our online approach is anytime, we vary the amount of time that is allocated to the algorithm for making a decision (referred to as runtime cutoff). Practitioners can vary this parameter to account for the maximum time they can afford to assign a request to a vehicle. Note that the passenger making the request need not wait for this duration to receive a feedback; whether a request can be serviced or not can be computed in less than a second using our approach (we only need to construct the RV graph to find at least one feasible action). 
%The runtime cutoff is the time it takes MC-VRP to find a nearly optimal decision, a process that can be optimized in the background. 
We consider two variants of our proposed MC-VRP approach. \textbf{\textit{MC-VRP (budget)}} uses our budget-based utility for scoring feasible actions in algorithm~\ref{algo: feasible actions}, while \textbf{\textit{MC-VRP (PTT)}} uses the PTT-based utility for scoring feasible actions. The MCTS evaluation as outlined in section~\ref{subsec: mcts evaluation} remains the same for both variants.

%Finally, we reiterate that As the time windows are hard constraints, a trip is only feasible if the passenger is picked up and dropped off between the early and late pickup windows.

% used to find the earliest time a trip can be picked up at the pickup location and the latest the passenger can be dropped off at the dropoff location. The early pickup window is the requested pickup time minus the time window. The late pickup window is the dropoff time plus the time window. 

\subsection{Baselines}\label{subsec: baselines}

We evaluate the performance of MC-VRP against the following baselines. For all baselines, we use the same number of vehicles (we present results by varying the number of vehicles) and vehicle capacity as MC-VRP. We compare our approach to two myopic online approaches (greedy and MA-RTV) and one non-myopic hybrid approach (DRLSA). %The baselines are described in detail as follows:
\pgfplotstableread[col sep=comma,]{data/drlsa/vehicle-3.csv}\drlsathree
\pgfplotstableread[col sep=comma,]{data/drlsa/vehicle-4.csv}\drlsafour
\pgfplotstableread[col sep=comma,]{data/drlsa/vehicle-5.csv}\drlsafive

\pgfplotstableread[col sep=comma,]{data/pnas/vehicles-3.csv}\pnasthree
\pgfplotstableread[col sep=comma,]{data/pnas/vehicles-4.csv}\pnasfour
\pgfplotstableread[col sep=comma,]{data/pnas/vehicles-5.csv}\pnasfive

\pgfplotstableread[col sep=comma,]{data/mcts/3_8_10_true_false_greedy_1000_20_1_Inf.csv}\greedythreeptt
\pgfplotstableread[col sep=comma,]{data/mcts/4_8_10_true_false_greedy_1000_20_1_Inf.csv}\greedyfourptt
\pgfplotstableread[col sep=comma,]{data/mcts/5_8_10_true_false_greedy_1000_20_1_Inf.csv}\greedyfiveptt

\pgfplotstableread[col sep=comma,]{data/mcts/3_8_10_true_false_mcts_1000_20_25_Inf.csv}\mctsthreeptt
\pgfplotstableread[col sep=comma,]{data/mcts/4_8_10_true_false_mcts_1000_20_25_Inf.csv}\mctsfourptt
\pgfplotstableread[col sep=comma,]{data/mcts/5_8_10_true_false_mcts_1000_20_25_Inf.csv}\mctsfiveptt

\pgfplotstableread[col sep=comma,]{data/mcts/3_8_10_true_true_greedy_1000_20_1_Inf.csv}\greedythreebudget
\pgfplotstableread[col sep=comma,]{data/mcts/4_8_10_true_true_greedy_1000_20_1_Inf.csv}\greedyfourbudget
\pgfplotstableread[col sep=comma,]{data/mcts/5_8_10_true_true_greedy_1000_20_1_Inf.csv}\greedyfivebudget

\pgfplotstableread[col sep=comma,]{data/mcts/3_8_10_true_true_mcts_1000_20_25_Inf.csv}\mctsthreebudget
\pgfplotstableread[col sep=comma,]{data/mcts/4_8_10_true_true_mcts_1000_20_25_Inf.csv}\mctsfourbudget
\pgfplotstableread[col sep=comma,]{data/mcts/5_8_10_true_true_mcts_1000_20_25_Inf.csv}\mctsfivebudget

\begin{figure*}[t]
\begin{tikzpicture}
\begin{axis}[
      boxplot/draw direction=y,
      xtick={1,2,3},
      xticklabels={{3 Vehicles}, {4 Vehicles}, {5 Vehicles}},
      width=\textwidth,
      height = 5cm,
      bugsResolvedStyle/.style={},
      ylabel={Service Rate},
      yticklabel=\pgfmathprintnumber{\tick}\,$\%$,
      xlabel={Number of Vehicles},
    ymin = 0, ymax = 110,
    ytick={0, 20, 40, 60, 80, 100},
    ymajorgrids={true},
    /pgfplots/boxplot/box extend=0.3,
       custom legend,
            legend style={at={(1,1)},anchor=south east,legend columns=6,
                column sep=0.5em}
    ]
    
\addlegendentry{DRLSA}
\addlegendentry{MA-RTV}
\addlegendentry{greedy-PTT}
\addlegendentry{greedy-budget}
\addlegendentry{MC-VRP (PTT)}
\addlegendentry{MC-VRP (budget)}

\addplot+[boxplot={box extend=0.1, draw position=0.65},drlsa, solid, fill=drlsa!20, mark=x] table [col sep=comma, y=servicerate] {\drlsathree};
\addplot+[boxplot={box extend=0.1, draw position=.79},pnas, solid, fill=pnas!20, mark=x] table [col sep=comma, y=servicerate] {\pnasthree};
\addplot+[boxplot={box extend=0.1, draw position=.93},greedyptt, solid, fill=greedyptt!20, mark=x] table [col sep=comma, y=servicerate] {\greedythreeptt};
\addplot+[boxplot={box extend=0.1, draw position=1.07},greedybudget, solid, fill=greedybudget!20, mark=x] table [col sep=comma, y=servicerate] {\greedythreebudget};
\addplot+[boxplot={box extend=0.1, draw position=1.21},mcvrpptt, solid, fill=mcvrpptt!20, mark=x] table [col sep=comma, y=servicerate] {\mctsthreeptt};
\addplot+[boxplot={box extend=0.1, draw position=1.34},mcvrpbudget, solid, fill=mcvrpbudget!20, mark=x] table [col sep=comma, y=servicerate] {\mctsthreebudget};

\addplot+[boxplot={box extend=0.1, draw position=1.65},drlsa, solid, fill=drlsa!20, mark=x] table [col sep=comma, y=servicerate] {\drlsafour};
\addplot+[boxplot={box extend=0.1, draw position=2.65},drlsa, solid, fill=drlsa!20, mark=x] table [col sep=comma, y=servicerate] {\drlsafive};

\addplot+[boxplot={box extend=0.1, draw position=1.79},pnas, solid, fill=pnas!20, mark=x] table [col sep=comma, y=servicerate] {\pnasfour};
\addplot+[boxplot={box extend=0.1, draw position=2.79},pnas, solid, fill=pnas!20, mark=x] table [col sep=comma, y=servicerate] {\pnasfive};

\addplot+[boxplot={box extend=0.1, draw position=1.93},greedyptt, solid, fill=greedyptt!20, mark=x] table [col sep=comma, y=servicerate] {\greedyfourptt};
\addplot+[boxplot={box extend=0.1, draw position=2.93},greedyptt, solid, fill=greedyptt!20, mark=x] table [col sep=comma, y=servicerate] {\greedyfiveptt};

\addplot+[boxplot={box extend=0.1, draw position=2.07},greedybudget, solid, fill=greedybudget!20, mark=x] table [col sep=comma, y=servicerate] {\greedyfourbudget};
\addplot+[boxplot={box extend=0.1, draw position=3.07},greedybudget, solid, fill=greedybudget!20, mark=x] table [col sep=comma, y=servicerate] {\greedyfivebudget};

\addplot+[boxplot={box extend=0.1, draw position=2.21},mcvrpptt, solid, fill=mcvrpptt!20, mark=x] table [col sep=comma, y=servicerate] {\mctsfourptt};
\addplot+[boxplot={box extend=0.1, draw position=3.21},mcvrpptt, solid, fill=mcvrpptt!20, mark=x] table [col sep=comma, y=servicerate] {\mctsfiveptt};

\addplot+[boxplot={box extend=0.1, draw position=2.34},mcvrpbudget, solid, fill=mcvrpbudget!20, mark=x] table [col sep=comma, y=servicerate] {\mctsfourbudget};
\addplot+[boxplot={box extend=0.1, draw position=3.34},mcvrpbudget, solid, fill=mcvrpbudget!20, mark=x] table [col sep=comma, y=servicerate] {\mctsfivebudget};

\end{axis}
\end{tikzpicture}
\caption{Service rate, defined as percentage of trips served, per day on 15 day test set for 3, 4 and 5 vehicles. 5 vehicles was enough to service all request on most days for MC-VRP (budget), MC-VRP (PTT) and MA-RTV. MC-VRP (budget) had the highest median service rate for 3 and 4 vehicles.
%of 87.0\% and 97.6\% respectively. 
The budget-based heuristic improves service rate for MC-VRP by 2.2\% and 2.1\% and greedy assignment by 3.4\% and 4.9\% for 3 and 4 vehicles respectively.}
\label{fig:service rate}
\end{figure*}
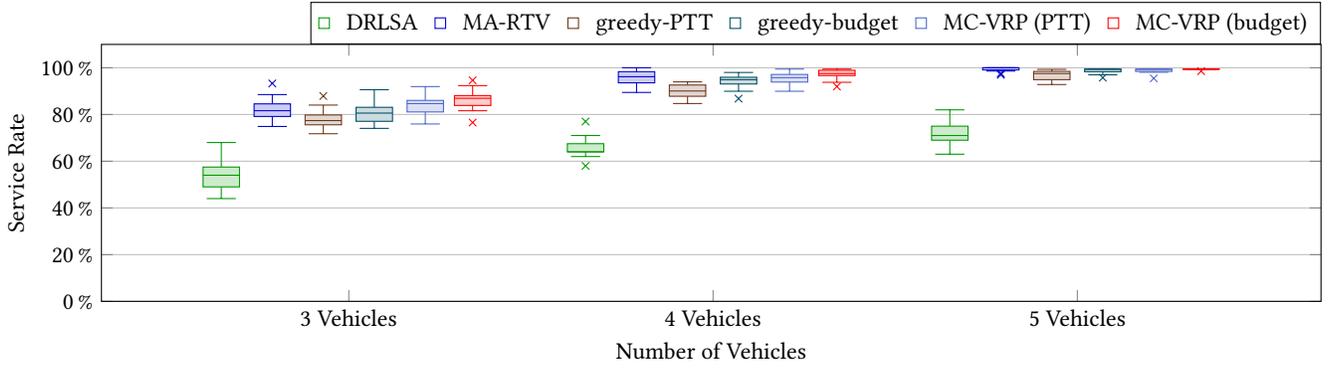
\begin{itemize}[leftmargin=*]
    \item 
    \textbf{Greedy assignment (greedy-PTT, greedy-budget):} Greedy assignment consists of first generating feasible actions according to algorithm~\ref{algo: feasible actions} and then selecting the action with the highest utility. For comparison, we include greedy assignment using both the PTT and budget utilities. We refer to greedy assignment with the PTT utility as greedy-PTT and greedy assignment with the budget utility as greedy-budget.
    %For a given request, greedy assignment chooses the set of route plans that maximizes the total utility of assigning the request to a vehicle and swaps between vehicles. \textcolor{red}{Ayan: this is not clear to me. Let us discuss this.}
    
    \item 
    \textbf{Deep reinforcement learning-based vehicle routing~\cite{joe2020deep} (DRLSA):}
    %\textbf{Deep reinforcement learning based vehicle routing~\cite{joe2020deep} (DRLSA):}
    \citeauthor{joe2020deep} combine deep reinforcement learning, which approximates a state-value function for routing, with a simulated-annealing based routing heuristic to solve the dynamic vehicle routing problem~\cite{van1987simulated}. The state representation of DRLSA is based on the total cost of the planned routes of the vehicles. While the original approach is designed for cargo-delivery problems, we add pickup and dropoff constraints given our problem setting. Also, the problem formulation by \citeauthor{joe2020deep}~\cite{joe2020deep} seeks to minimize the sum of the travel and waiting times of all vehicles. It serves all requests and incurs a penalty cost for time-window violations.
    To apply DRLSA to our problem formulation, where time-windows constraints are strict, we take the route plans output by DRLSA and iteratively remove requests that violate time windows until a maximal feasible set of requests is left.
    Further, for a fair comparison, we run the DRLSA algorithm itself with shorter time windows but remove requests that violate the original time windows, which we found significantly improves the service rate of DRLSA in our setting.
    %To reduce the number of requests for time window constraints violation with a small margin, we consider early late-dropoff-window to calculate the penalty, but the original late-dropoff-window to calculate the service rate. 
    %For a fair comparison, we filter out requests with the time window constraints violation in such a way so that we can have the maximum service rate. 
    Finally, we point out that since there was no open-source implementation available for this approach, we implemented DRLSA for this paper.
    %\Salah{Ayan: Salah, you wrote that you added passenger constraints, did icaps have some capacity constraint as well? Salah: No based on the algorithm, they pickup all the items and then deliver}
    
    % We get an average service rate of 71.73\% for five vehicles, 65.73\% for four vehicles, and 53.73\% for three vehicles.
    % The original problem is a customer's orders delivery problem. Although our problem setting is somewhat different from that, we make it work by adding additional constraints but following a similar approach. We needed to add the pickup-dropoff and the number of passengers constraints.  
    % Although, in this experiment, the objective is to maximize the service rate (a request qualifies if it can satisfy the time window constraints) where this algorithm minimizes the sum of total travel and waiting times of all vehicles and penalty cost for time window violations. To reduce the number of requests of time window constraint violation with a small margin, we consider early late-dropoff-window to calculate the penalty, but the original late-dropoff-window to calculate the service rate. To count the service rate, we filter out time window violated requests and update the travel time for the remaining routes (\textcolor{red}{might need to write this line in a better way}). Note that the original implementation is not available on the internet, so we had to implement it ourselves. 
    
    \item 
    \textbf{Myopic and Anytime trip-vehicle assignment-based on RTV graphs~\cite{alonso2017demand} (MA-RTV):}
    %\textbf{Myopic and Anytime trip-vehicle assignment based on RTV graphs~\cite{alonso2017demand} (MA-RTV):} 
    % We use OpenRidepoolSimulator\footnote{https://github.com/MAS-Research/OpenRidepoolSimulator} to design the experimental setup for this approach. The base implementation closely follows the design presented by \citeauthor{alonso2017demand}~\cite{alonso2017demand} and is publicly available. 
    We also compare our approach with an anytime algorithm that batches requests together and then maximizes the myopic utility for the specific batch of requests. For each batch, the algorithm creates an RTV graph by checking \textit{shareability} between requests as well as requests and vehicles. Since the best passengers-vehicle pair is found in each batch, the approach works well in practice despite being myopic (our experiments confirm this). We refer to this baseline as \textit{MA-RTV}. To adapt MA-RTV for our problem setting, we set the parameters as follows. First, as MA-RTV is not able to handle requests in advance, we set the request arrival time to the early time window. Second, we set the latest dropoff time to 30 minutes (15 minutes for early time window and 15 minutes for late time window) plus the shortest path between the pickup and dropoff locations. Therefore, the time window requirements are equivalent to our problem setup. Lastly, MA-RTV is a batching approach which waits for a set period of time before grouping requests together for assignment. In the paratransit setting, requests rarely arrive very close together and are expected to be handled one at a time, typically over the phone. Therefore we set the batch interval to 20 seconds, which closely mimics the observed paratransit data.
    %As the MA-RTV approach was designed to respond to requests as soon as they arrive, we feed the requests to the algorithm at the beginning of the requested pickup window. 
    %To calculate the dropoff window, we use $\dots$ \textcolor{red}{Ayan: this is not immediately clear to me.} To replicate the sparsity of requests in the paratransit setting (in practice, multiple requests rarely arrive very close to each other), we set batch interval to 20 seconds. 
    
    % that implement the design presented in the paper, which is publicly available. We needed to translate our parameters to the inputs of the simulators. Since the simulation does not have time window, we subtracted 15 minutes from the time that requests coming in to reflect the start of the time window. We set maximum waiting time and maximum detour of the customer to be 30 minutes so that the length of the time window to be 30 minutes. We used the same number of vehicles and vehicle capacity. Batch interval is set to 20 seconds. 
\end{itemize}

%\textbf{Parameter Tuning:} For all baselines, we used 3 days of paratransit data  for hyperparameter tuning. We present details about hyperparameter search in the technical appendix (section~\ref{sec_appendix:hyper}). 

\subsection{Parameter Tuning}\label{sec_appendix:hyper}

% \usepackage{graphicx}
\begin{table}[t]
 \footnotesize
\centering
\caption{Parameter settings}
\begin{tabular}{|l|l|}
\hline
\textbf{Parameter} & \textbf{Values} \\ \hline
Number of vehicles ($M$) & 3, 4, 5 \\ \hline
Vehicle capacity & 8 \\ \hline
Request arrival time & 60m \\ \hline
Time window & 15m \\ \hline
Runtime cutoff & Inf, 30s \\ \hline
$K_{max}$ & 10 \\ \hline
MCTS depth & 20 \\ \hline
MCTS iterations &  1000 \\ \hline
$n_{chains}$ &  25 \\ \hline
\end{tabular}%
\label{tab: parameters}
\end{table}

We used 3 days of paratransit data  for hyperparameter tuning. MC-VRP has two parameters for tuning---the depth of the search tree and the number of feasible actions. MCTS tree depth, which is the number of future requests to consider from the generative model, was varied between $\{10,20,30\}$. The number of feasible actions to explore, denoted by $K_{max}$, effectively sets the maximum branching factor for MCTS. We varied this parameter between $\{10, 15, 20, 25, 30\}$. We performed a grid search using the calibration set and found the best parameters in terms of service rate (MCTS depth of 20 and $K_{max}=10$). The final parameters for MC-VRP are provided in Table~\ref{tab: parameters}.

The calibration set was also used to select the parameters for DRLSA. The neural network of the DRLSA baseline approach consists of 1 hidden layer with 64 neurons. The activation function of the hidden layer is rectified linear (ReLu) and linear for the output layer. We trained the network with a batch size of 32, the discount factor for future reward is 0.99, and the learning rate is 0.01. We found the best result with $K_{max}=500$ for Simulated Annealing. Similar to MC-VRP, we used grid search to find the best parameters for DRSLA.

\subsection{Reproducibility}
Our source code, implementation of the baseline approaches, and samples of our datasets are available online (\url{https://github.com/smarttransit-ai/iccps-2022-paratransit-public}). We implemented MC-VRP in the Julia programming language using the pomdps.jl framework~\cite{egorov2017pomdps}. Results presented in this paper were obtained using the Chameleon testbed supported by the National Science Foundation \cite{keahey2020lessons}.

% hidden layer 64, relu
% output layer 1, linear
% buffer_capacity=50000,
% batch_size=32
        % self.GAMMA = 0.99                 # discount factor
        % self.LEARNING_RATE = 0.01
        % self.C = 1                        # number of steps to update target value function
        % self.EPSILON = 0.1                # select a random decision

% \textcolor{red}{MW: Salah we should probably put something more here for DRLSA, at least what the final NN configuration was for the model.}. 

\subsection{Results}

\textbf{Service Rates:} Our primary objective is to maximize the number of requests serviced each day. The service rate per day for the 15 days in the evaluation (test) set is provided in figure~\ref{fig:service rate} with fleet sizes of 3, 4, and 5 vehicles. We first present results in a setting where MC-VRP (budget) and MC-VRP (PTT) are allowed to run for 1000 iterations without early termination. First, we note that 5 vehicles is enough to service all of the requests on most days as MC-VRP (budget), MC-VRP (PTT), and MA-RTV all have a median service rate of 100\%. We observe that MC-VRP (budget) outperforms all baselines for fleet sizes of 3 and 4 with with a median service rate of 87.0\% and 97.6\% respectively. MC-VRP (PTT) had a service rate of 84.7\% and MA-RTV had a service rate of 81.7\% for 3 vehicles. MA-RTV outperformed MC-VRP (PTT) in the case of 4 vehicles with a median service rate of 96.2\% compared to 95.8\%.

We also observe that the budget-based heuristic works better than the travel time-based heuristic for both Monte Carlo tree search as well as greedy assignment. Indeed, MC-VRP (budget) results in a higher service rate than MC-VRP (PTT) and greedy-budget outperforms greedy-PTT for all fleet sizes. This indicates that the budget utility, which aims to maximize time for which a vehicle has no passengers on-board, can be used to quickly compute promising actions to explore. It is important to note that while we focus on paratransit services, the budget-based heuristic can be applied for other DVRPs with capacity and time window constraints. 
% out-performs PTT as a heuristic for scoring route plans in the context of servicing future requests in our setting. 
We also observe that DRLSA had the lowest service rate across all fleet sizes. An influencing factor, as discussed in section~\ref{subsec: baselines}, is that time windows are soft constraints in DRSLA which is not particularly suited to paratransit settings.

\textbf{Computation Time:} The computation time per request for MC-VRP and the baselines is shown in figure~\ref{fig:computation time}. MC-VRP (budget) takes slightly longer than MC-VRP (PTT) to compute a decision, with a median computation time of 38 seconds, 49 seconds, and 40 seconds as compared to 36 seconds, 44 seconds, and 37 seconds for 3, 4, and 5 vehicles respectively. DRLSA, MA-RTV, and greedy all had median computation times less than 3 seconds. In this context, the fact that MC-VRP is both non-myopic and fully online is reflected in the higher runtimes; while the observed runtimes are acceptable in our setting, i.e., paratransit services, the application of online and non-myopic methods remains an open question in general VRP settings. Note that if the MCTS evaluation of our approach is not used (i.e., the early stopping time approaches 0), our approach simplifies to the greedy baseline, which takes less than a second on average. This observation means that the majority of computation time in our approach is spent on MCTS.

%However, we point out that when a passenger makes a request, they do not necessarily have to wait for MC-VRP to compute a nearly optimal decision. In practice, the passenger can be provided a confirmation based on a computationally cheap policy, e.g., greedy assignment. The MCTS evaluation can be run in the background between calls. 

As discussed, the MCTS evaluation can be run in the background between calls. Therefore, a potentially limiting factor for our approach is the rate at which requests arrive; even when running in background between requests, MC-VRP must be stopped on the arrival of a new request. Since MC-VRP is an \textit{anytime} algorithm, it is possible to set a maximum computation time per request. In such a setting, the algorithm is stopped early (in our case, before the default number of iterations is reached), but outputs a feasible solution. In our dataset, 99\% of requests arrive more than 30 seconds after the previous request and the median time between requests was approximately 5 minutes. Therefore, we evaluate the performance of MC-VRP with a 30 second cut-off time in figure~\ref{fig:runtime}.
%We present the percent change in service rate for each day in the test set (with respect to MCTS being allowed to finish 1000 iterations) is shown in figure~\ref{fig:runtime}. 
We observe only a negligible decrease in service rates; the median service rate decreases by 1.2\% for 3 vehicles and 0\% for 4 and 5 vehicles.

%\textcolor{blue}{Note that if the MCTS evaluation of our approach is not used (e.g. the early stopping time approaches 0), our approach simplifies to the greedy baseline. However, in our dataset, the average time between requests is approximately 5 minutes and 99\% of requests arrive more than 30 seconds after the previous request. 
%To test MC-VRP in a setting where it is stopped as soon as the next request arrives, we ran MC-VRP (budget) on the same test set with a maximum computation time of 30 seconds.}

%In our dataset, 99\% of requests arrive more than 30 seconds after the previous request and the median time between requests was approximately 5 minutes. Therefore, we evaluate the performance of MC-VRP with a 30 second cut-off time.

%As MC-VRP is fully online, the time per request is higher than DRLSA, MA-RTV and greedy. 

%Reference numbers (service rate):
%\begin{itemize}
%    \item PNAS
%    \begin{itemize}
%        \item median: 81.6216, 96.1783, 100.0000
%        \item mean: 82.233713, 95.689647, 99.385500
%    \end{itemize}
%    \item DRLSA
%    \begin{itemize}
%        \item median: 54.0, 64.0, 71.0
%        \item mean: 53.733333, 65.733333, 71.533333
%    \end{itemize}
%    \item MC-VRP (budget)
%    \begin{itemize}
%        \item median: 87.0 97.54, 100
%        \item mean: 86.45, 97.29, 99.41
%    \end{itemize}
%    \item MC-VRP (PTT)
%    \begin{itemize}
%        \item median: 84.7, 95.8, 100
%        \item mean: 
%    \end{itemize}
%\end{itemize}

%\include{data/servicerate}
\pgfplotstableread[col sep=comma,]{data/drlsa/vehicle-3.csv}\drlsathree
\pgfplotstableread[col sep=comma,]{data/drlsa/vehicle-4.csv}\drlsafour
\pgfplotstableread[col sep=comma,]{data/drlsa/vehicle-5.csv}\drlsafive

\pgfplotstableread[col sep=comma,]{data/pnas/vehicles-3.csv}\pnasthree
\pgfplotstableread[col sep=comma,]{data/pnas/vehicles-4.csv}\pnasfour
\pgfplotstableread[col sep=comma,]{data/pnas/vehicles-5.csv}\pnasfive

\pgfplotstableread[col sep=comma,]{data/mcts/3_8_10_true_false_greedy_1000_20_1_Inf.csv}\greedythreeptt
\pgfplotstableread[col sep=comma,]{data/mcts/4_8_10_true_false_greedy_1000_20_1_Inf.csv}\greedyfourptt
\pgfplotstableread[col sep=comma,]{data/mcts/5_8_10_true_false_greedy_1000_20_1_Inf.csv}\greedyfiveptt

\pgfplotstableread[col sep=comma,]{data/mcts/3_8_10_true_false_mcts_1000_20_25_Inf.csv}\mctsthreeptt
\pgfplotstableread[col sep=comma,]{data/mcts/4_8_10_true_false_mcts_1000_20_25_Inf.csv}\mctsfourptt
\pgfplotstableread[col sep=comma,]{data/mcts/5_8_10_true_false_mcts_1000_20_25_Inf.csv}\mctsfiveptt

\pgfplotstableread[col sep=comma,]{data/mcts/3_8_10_true_true_greedy_1000_20_1_Inf.csv}\greedythreebudget
\pgfplotstableread[col sep=comma,]{data/mcts/4_8_10_true_true_greedy_1000_20_1_Inf.csv}\greedyfourbudget
\pgfplotstableread[col sep=comma,]{data/mcts/5_8_10_true_true_greedy_1000_20_1_Inf.csv}\greedyfivebudget

\pgfplotstableread[col sep=comma,]{data/mcts/3_8_10_true_true_mcts_1000_20_25_Inf.csv}\mctsthreebudget
\pgfplotstableread[col sep=comma,]{data/mcts/4_8_10_true_true_mcts_1000_20_25_Inf.csv}\mctsfourbudget
\pgfplotstableread[col sep=comma,]{data/mcts/5_8_10_true_true_mcts_1000_20_25_Inf.csv}\mctsfivebudget

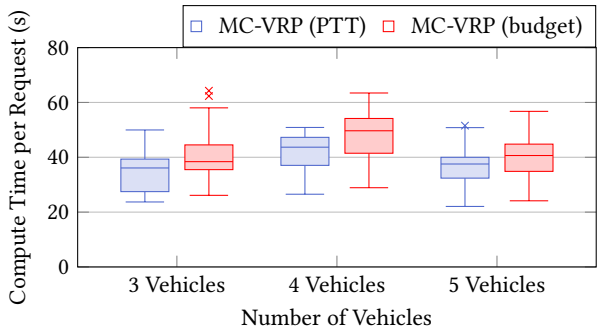
\begin{figure}[t]
\begin{tikzpicture}
\begin{axis}[
      boxplot/draw direction=y,
      xtick={1,2,3},
      xticklabels={{3 Vehicles}, {4 Vehicles}, {5 Vehicles}},
      width=\columnwidth,
      height = 4.5cm,
      bugsResolvedStyle/.style={},
      ylabel={Compute Time per Request (s)},
      %yticklabel=\pgfmathprintnumber{\tick}\,$\%$,
      xlabel={Number of Vehicles},
    ymin = 0, ymax = 80,
    ytick={0, 20, 40, 60, 80},
    ymajorgrids={true},
    /pgfplots/boxplot/box extend=0.3,
       custom legend,
            legend style={at={(1,1)},anchor=south east,legend columns=2,
                column sep=0.5em}
    ]
    
\addlegendentry{MC-VRP (PTT)}
\addlegendentry{MC-VRP (budget)}
    
\addplot+[boxplot={box extend=0.3, draw position=0.8},mcvrpptt, solid, fill=mcvrpptt!20, mark=x] table [col sep=comma, y=timeperrequest] {\mctsthreeptt};
\addplot+[boxplot={box extend=0.3, draw position=1.2},mcvrpbudget, solid, fill=mcvrpbudget!20, mark=x] table [col sep=comma, y=timeperrequest] {\mctsthreebudget};

\addplot+[boxplot={box extend=0.3, draw position=1.8},mcvrpptt, solid, fill=mcvrpptt!20, mark=x] table [col sep=comma, y=timeperrequest] {\mctsfourptt};
\addplot+[boxplot={box extend=0.3, draw position=2.2},mcvrpbudget, solid, fill=mcvrpbudget!20, mark=x] table [col sep=comma, y=timeperrequest] {\mctsfourbudget};

\addplot+[boxplot={box extend=0.3, draw position=2.8},mcvrpptt, solid, fill=mcvrpptt!20, mark=x] table [col sep=comma, y=timeperrequest] {\mctsfiveptt};
\addplot+[boxplot={box extend=0.3, draw position=3.2},mcvrpbudget, solid, fill=mcvrpbudget!20, mark=x] table [col sep=comma, y=timeperrequest] {\mctsfivebudget};

\end{axis}
\end{tikzpicture}
\vspace{-0.1in}
\caption{Computation time per request in seconds for each day in the test set for 3, 4 and 5 vehicles. Median computation time per request for MC-VRP (budget) is 38 seconds, 49 seconds and 40 seconds for 3, 4 and 5 vehicles respectively. DRLSA, MA-RTV and greedy all had median computation times less than 3 seconds per request.}
\label{fig:computation time}
\end{figure}
\pgfplotstableread[col sep=comma,]{data/runtime/runtime.csv}\runtime

\begin{figure}[t]
\begin{tikzpicture}
\begin{axis}[
      boxplot/draw direction=y,
      xtick={1,2,3},
      xticklabels={{3 Vehicles}, {4 Vehicles}, {5 Vehicles}},
      width=\columnwidth,
      height = 4.5cm,
      bugsResolvedStyle/.style={},
      ylabel={\% Change in Service Rate},
      yticklabel=\pgfmathprintnumber{\tick}\,$\%$,
      xlabel={Number of Vehicles},
    ymin = -6, ymax = 6,
    ytick={-6, -4, -2, 0, 2, 4, 6},
    ymajorgrids={true},
    /pgfplots/boxplot/box extend=0.3,
       custom legend,
            legend style={at={(1,1)},anchor=south east,legend columns=2,
                column sep=0.5em}
    ]
    
\addlegendentry{MC-VRP (budget)}

\addplot+[boxplot={box extend=0.4, draw position=1},mcvrpbudget, solid, fill=mcvrpbudget!20, mark=x] table [col sep=comma, y=threevehicles] {\runtime};
\addplot+[boxplot={box extend=0.4, draw position=2},mcvrpbudget, solid, fill=mcvrpbudget!20, mark=x] table [col sep=comma, y=fourvehicles] {\runtime};
\addplot+[boxplot={box extend=0.4, draw position=3},mcvrpbudget, solid, fill=mcvrpbudget!20, mark=x] table [col sep=comma, y=fivevehicles] {\runtime};

\end{axis}
\end{tikzpicture}
\vspace{-0.1in}
\caption{Runtime analysis: Percent change in service rate for MC-VRP (budget) with a cutoff of 30 seconds per request compared to no cutoff. For three vehicles there was a median of 1.2 percent decrease in service rate on test set while the computation cutoff had negligible affect for four and five vehicles.}
\label{fig:runtime}
\end{figure}
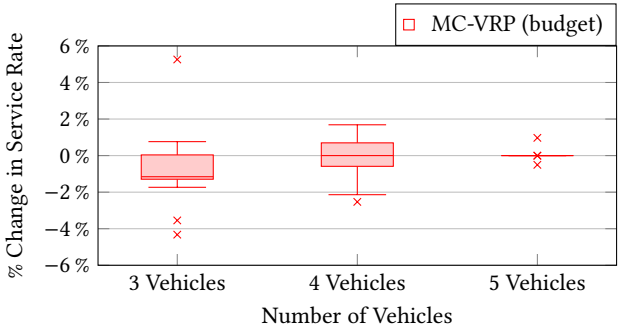

%\begin{figure*}[t]
%\includegraphics[width=\textwidth]{images/servicerate.pdf}
%\caption{Service rate per day on 15 day test set for 3, 4 and 5 vehicles. MC-VRP with budget heuristic outperforms DRLSA and RTV. The budget heuristic improves both MC-VRP and greedy. \textcolor{red}{MW: put median values in caption, expand discussion, will redo this plot in latex tomorrow.}}
%\label{fig:service rate}
%\end{figure*}

%\begin{figure*}[t]
%\includegraphics[width=\textwidth]{images/timeperrequest.pdf}
%\caption{Computation time per request on 15 day test set for 3, 4 and 5 vehicles. \textcolor{red}{MW: put median values in caption, expand discussion, redo plot in latex}}
%\label{fig:computation time}
%\end{figure*}

%\begin{figure}[t]
%\includegraphics[width=.9\columnwidth]{images/runtime.pdf}
%\caption{Runtime analysis: Percent change in service rate for MC-VRP with budget heuristic with a cutoff of 30 seconds per request compared to no cutoff. For three vehicles there was a median of 1.2 percent decrease in service rate on test set while the computation cutoff had negligible affect for four and five vehicles. \textcolor{red}{MW: redo plot in latex}}
%\label{fig:runtime}
%\end{figure}

\textbf{Robustness:} To evaluate the robustness of the proposed approach with respect to changing environmental conditions, we change the travel time distribution in the city. For experiments, we assume the existence of a service that uses short term observations about an environmental variable (e.g., travel times) and provides an updated model. In practice, transit agencies can use services like Google maps for this purpose. In this case, when a user requests a ride, we can use the updated travel time matrix to give an accurate estimate of the dropoff time. We assume that all approaches have access to the updated travel time matrix at inference time, i.e. when a user requests a ride. However, approaches that rely on models trained offline do not have the ability to update their model. 

\begin{figure}[t]
     \centering
   \includegraphics[width=\columnwidth]{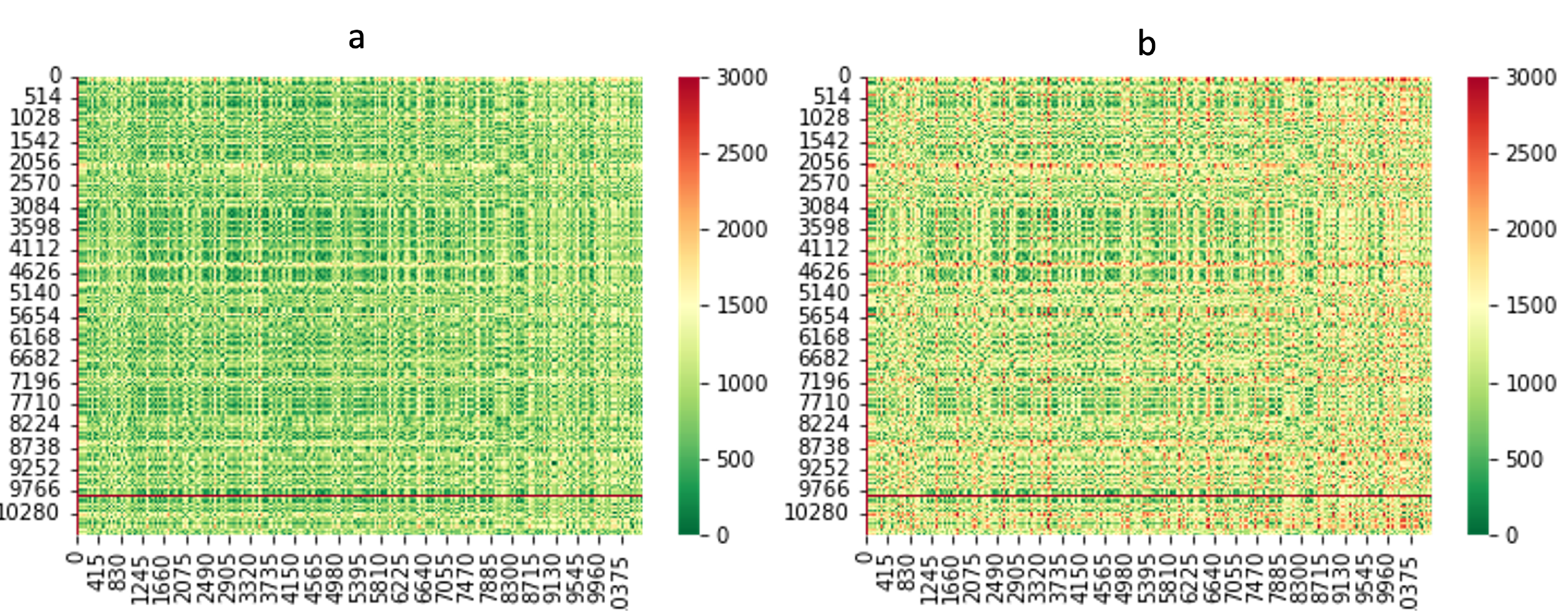}
          \caption{(a) Free-flow speed vs (b) Irregular travel times in seconds. Irregular travel times represent a congested roadway network. Each cell shows the travel time from a location x to another location y. There are total 10788x10788 combinations in the travel time matrix.}
        \label{fig:traveltimes}
\end{figure}

We generate a modified travel time matrix that resembles a congested road network. The distribution of the free-flow travel time matrix and the congested travel time matrix is shown in Figure~\ref{fig:traveltimes}. The original travel time matrix was calculated by finding shortest paths between all nodes in an OSM graph where the edge weights were travel time between the nodes using free flow speed. Each edge in the OSM graph had a unique OSM ID which mapped to roadway speeds in the INRIX dataset. The INRIX dataset included average roadway speeds per hour for each day in the week. For each OSM ID we took the speed at the 5th percentile and updated the edge weights accordingly. On average, the congested travel time matrix had speeds that were 30\% slower than the free-flow speed. %We describe the process to generate the updated matrix in the technical appendix (section~\ref{sec_appendix:travel_time}).

%\subsection{Congested Travel Time Matrix}\label{sec_appendix:travel_time}

\pgfplotstableread[col sep=comma,]{data/robust/robust.csv}\robust
\pgfplotstableread[col sep=comma,]{data/robust/robustgreedy.csv}\robustgreedy

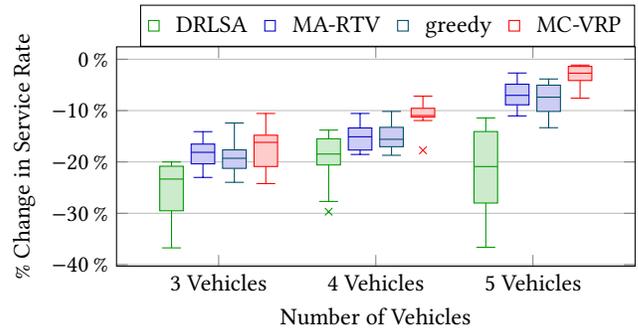
\begin{figure}[t]
\begin{tikzpicture}
\begin{axis}[
      boxplot/draw direction=y,
      xtick={1,2,3},
      xticklabels={{3 Vehicles}, {4 Vehicles}, {5 Vehicles}},
      width=\columnwidth,
      height = 4.5cm,
      bugsResolvedStyle/.style={},
      ylabel={\% Change in Service Rate},
      yticklabel=\pgfmathprintnumber{\tick}\,$\%$,
      xlabel={Number of Vehicles},
    ymajorgrids={true},
    /pgfplots/boxplot/box extend=0.3,
       custom legend,
            legend style={at={(1,1)},anchor=south east,legend columns=4,
                column sep=0.5em}
    ]
    
\addlegendentry{DRLSA}
\addlegendentry{MA-RTV}
\addlegendentry{greedy}
\addlegendentry{MC-VRP}
    
\addplot+[boxplot={box extend=0.15, draw position=0.7},drlsa, solid, fill=drlsa!20, mark=x] table [col sep=comma, y=drlsathree] {\robust};
\addplot+[boxplot={box extend=0.15, draw position=.9},pnas, solid, fill=pnas!20, mark=x] table [col sep=comma, y=rtvthree] {\robust};
\addplot+[boxplot={box extend=0.15, draw position=1.1},greedybudget, solid, fill=pnas!20, mark=x] table [col sep=comma, y=greedyvrpthree] {\robustgreedy};
\addplot+[boxplot={box extend=0.15, draw position=1.3},mcvrpbudget, solid, fill=mcvrpbudget!20, mark=x] table [col sep=comma, y=mcvrpthree] {\robust};

\addplot+[boxplot={box extend=0.15, draw position=1.7},drlsa, solid, fill=drlsa!20, mark=x] table [col sep=comma, y=drlsafour] {\robust};
\addplot+[boxplot={box extend=0.15, draw position=1.9},pnas, solid, fill=pnas!20, mark=x] table [col sep=comma, y=rtvfour] {\robust};
\addplot+[boxplot={box extend=0.15, draw position=2.1},greedybudget, solid, fill=pnas!20, mark=x] table [col sep=comma, y=greedyvrpfour] {\robustgreedy};
\addplot+[boxplot={box extend=0.15, draw position=2.3},mcvrpbudget, solid, fill=mcvrpbudget!20, mark=x] table [col sep=comma, y=mcvrpfour] {\robust};

\addplot+[boxplot={box extend=0.15, draw position=2.7},drlsa, solid, fill=drlsa!20, mark=x] table [col sep=comma, y=drlsafive] {\robust};
\addplot+[boxplot={box extend=0.15, draw position=2.9},pnas, solid, fill=pnas!20, mark=x] table [col sep=comma, y=rtvfive] {\robust};
\addplot+[boxplot={box extend=0.15, draw position=3.1},greedybudget, solid, fill=pnas!20, mark=x] table [col sep=comma, y=greedyvrpfive] {\robustgreedy};
\addplot+[boxplot={box extend=0.15, draw position=3.3},mcvrpbudget, solid, fill=mcvrpbudget!20, mark=x] table [col sep=comma, y=mcvrpfive] {\robust};

\end{axis}
\end{tikzpicture}
\vspace{-0.1in}
\caption{Robustness to roadway congestion: Percent change in service rate for DRLSA, MA-RTV, greedy-budget and MC-VRP (budget) on test set using congested travel time matrix compared to free-flow speed matrix. MC-VRP (budget) had less of a decrease in service rate for 3, 4 and 5 vehicle fleets compared to the baselines.}
\label{fig:robust}
\end{figure}

%To evaluate the robustness of our approach to changes in traffic conditions we generated a second travel time matrix representing a congested road network. The original travel time matrix was calculated by finding shortest paths between all nodes in an OSM graph where the edge weights were travel time between the nodes using free flow speed. Each edge in the OSM graph had a unique OSM ID which mapped to roadway speeds in the INRIX dataset. The INRIX dataset included average roadway speeds per hour for each day in the week. For each OSM ID we took the speed at the 5th percentile and updated the edge weights accordingly. The distribution of the free-flow travel time matrix and the congested travel time matrix is shown in Figure~\ref{fig:traveltimes}. On average, the congested travel time matrix had speeds that were 30\% slower than the free-flow speed.

We evaluated DRLSA, MA-RTV, greedy-budget and MC-VRP (budget) on the 15 day evaluation set using the congested travel time network and compared the resulting service rates to those generated without congestion. We observe that MC-VRP (budget) incurs less of a decrease in service rate compared to DRLSA, MA-RTV and greedy-budget across all sizes of vehicle fleets as shown in figure~\ref{fig:robust}. As expected, DRLSA had the greatest decreases in service rate due to its reliance on a value function that was trained offline using the free-flow travel time matrix. While hybrid approaches such as DRLSA have access to the real-time travel conditions, it is difficult to re-train offline components, resulting in a higher degradation in performance. Both MA-RTV and greedy-budget outperform DRLSA when evaluated for robustness. This result is expected since MA-RTV and greedy-budget are both online approaches, it is able to adapt to the updated matrix. We also observe that MC-VRP (budget) performs better MA-RTV and greedy-budget, showing that our non-myopic online solution improved upon the myopic online approaches as well.

\section{Related Work}

Vehicle Routing Problems (VRPs) can be broadly classified as either static or dynamic~\cite{psaraftis2016dynamic}. In static VRPs, all inputs are received before optimizing routes, whereas in dynamic VRPs inputs are updated concurrently with the determination of the route. The focus of this paper is on dynamic VRPs (DVRP), which can be either dynamic-deterministic, in which no stochastic information about future inputs is known, or dynamic-stochastic, in which some probabilistic information is known about the inputs that dynamically evolve~\cite{pillac2013review}. These stochastic inputs can include models over travel times, demands, and customer information~\cite{ritzinger2016survey}. DVRPs can be solved to maximize myopic rewards~\cite{alonso2017demand} or a non-myopic  utility function~\cite{joe2020deep,shah2020neural}. Exact methods for solving DVRPs seek to find an optimal solution, but are often constrained to small problem instances due to computational complexity, e.g., such approaches include column-generation~\cite{chen2006real} and the set-partitioning method~\cite{christiansen2007branch}. Metaheuristic approaches have also been applied to DVRPs, including particle swarm optimization~\cite{khouadjia2012comparative}, genetic algorithms~\cite{taniguchi2004intelligent}, and tabu search~\cite{gendreau1999parallel, attanasio2004parallel}. Decision theoretic approaches have also been applied to DVRPs. DVRP is a sequential decision-making problem, and can be modeled as a Markov Decision Process (MDP). One approach is to use a conventional MDP structure, where actions consist of determining the next customer to serve at each decision epoch. Another approach, proposed by \citeauthor{ulmer2017route}~\cite{ulmer2017route}, is based on using a route-based MDP where the state-action space includes not only assigning an incoming request to a vehicle, but also optimizing the vehicles' routes. 

There are three ways to solve the DVRP MDPs: 
%Three ways to approach solving DVRPs formulated as MDPs are 
offline, online, and hybrid solution methods. Offline methods pre-compute a policy that is queried while executing a plan~\cite{shah2020neural,ulmer2019preemptive,nazari2018reinforcement}.
% One challenge with learning such a policy is overcoming the curse of dimensionality~\cite{pillac2013review}. Offline methods that have been applied to DVRP include approximate dynamic programming (ADP)~\cite{shah2020neural}, approximate value iteration (AVI)~\cite{ulmer2019preemptive}, and reinforcement learning~\cite{nazari2018reinforcement}. 
Offline policies can be slow to learn, but can make decisions very quickly at execution time, and are therefore useful when there are strict time constraints on decision-making. Online solution methods perform computations during plan execution. These are generally sampling approaches, and only focus on the states of the system relevant to the current decision being made. Online methods include rollout algorithms~\cite{secomandi2001rollout, goodson2017rollout} and multiple scenario approach (MSA)~\cite{bent2004scenario}. Online approaches have typically been applied to problem settings without strict time constraints on decision-making (the time it takes to compute a decision), and in dynamic environments where policies generated offline can become stale. Hybrid solution methods attempt to combine offline and online approaches to leverage the strengths of both. For example, \citeauthor{ulmer2019offline}~\cite{ulmer2019offline} propose an approach that embeds a value function learned using ADP into an online rollout algorithm. \citeauthor{joe2020deep}~\cite{joe2020deep} propose a similar approach, which combines Deep Reinforcement Learning to approximate a value function offline with an online simulated annealing approach.

\section{Conclusion}

We design a non-myopic online approach for DVRP for paratransit services that is robust to environmental changes by construction and scalable to real-world applications. 
%Our approach, based on Monte Carlo tree search, can compute a nearly-optimal action for a given state efficiently. 
To tackle the intractable action space, we leverage the structure of the problem to design heuristics that can sample promising actions for evaluation through MCTS. Our experimental results demonstrate superior performance and increases robustness against state-of-the-art baselines.

\section*{Acknowledgements}
This material is based upon work sponsored by National Science Foundation under grant CNS-1952011 and Department of Energy under Award Number DE-EE0009212.

\balance
\bibliographystyle{ACM-Reference-Format} 
\bibliography{main}
\clearpage
%\newpage
%\appendix
%\input{appendix_1}
\end{document}